# Mobility Enhancement for Elderly

**RAMVIYAS N P**
**2008JID2945**

Guided by

**Prof. D.T.SHAHANI**
IIT Delhi

**Prof. PASCAL FROSSARD**
EPFL

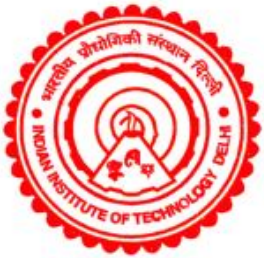
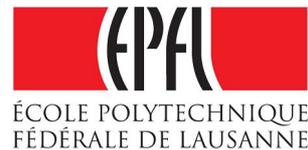

**Instrument Design Development Center**
**Indian Institute of Technology Delhi**
Hauz Khas, New Delhi 110016

Master Thesis Project
June 2010

# CERTIFICATE

This is to certify that the project entitled **"Mobility Enhancement for the Elderly"** submitted to the Instrument Design Development Centre, Indian Institute of Technology Delhi by **RAMVIYAS N P** in partial fulfillment for the requirement of the award of the degree of "Master of Technology" in "Instrument Technology" is a record of bonfire work carried out by him under my supervision and guidance.

Further to the best of my knowledge, this report has not been submitted for any other degree or diploma.

Date: 25<sup>th</sup> June 2010                               **Prof. D.T.Shahani**

Place: New Delhi                                                Professor,

                                                                                     IDDC, IIT Delhi

# ACKNOWLEDGEMENT

I express my profound sense of gratitude to my supervisors **Prof. D.T.Shahani (IITD) and Prof. Pascal Frossard (EPFL)** for their immense interest, invaluable guidance, moral support and constant encouragement during the tenure of this work.

My deepest thanks to all the professors of IDDC, for their invaluable teaching and guidance in all ways of my life.

I am also indebted to Mr. Masood Ali and Mr. M. S. Negi of MDIT Lab for their coordination, encouragement at every stage of this project.

I express my thanks to Mr. Vijayaragavan Thirumalai, Ms.Teodora Kostic, students of the Signal Processing Lab in EPFL and Mr.Satish, Mr.Ramkumar, students of the MDIT Lab in IITD for their wonderful support and help they offered in realizing the goal of the project

I would also like to thank all my classmates for their support in my thesis work.

Finally, I am grateful to my parents, without whom none of this world is possible. They are an unbelievable inspiration to me for instilling the love of learning in me and for teaching me about integrity, dignity and respect.

<div align="right">

RAMVIYAS N P
2008JID2945
M. Tech (Instrument Technology)

</div>

# Abstract

Loss of Mobility is a common handicap to senior citizens. It denies them the ease of movement they would like to have like outdoor visits, movement in hospitals, social outgoings, but more seriously in the day to day in-house routine functions necessary for living etc.

Trying to overcome this handicap by means of servant or domestic help and simple wheel chairs is not only costly in the long run, but forces the senior citizen to be at the mercy of sincerity of domestic helps and also the consequent loss of dignity. In order to give a dignified life, the mobility obtained must be at the complete discretion, will and control of the senior citizen. This can be provided only by a reasonably sophisticated and versatile wheel chair, giving enhanced ability of vision, hearing through man-machine interface, and sensor aided navigation and control. More often than not senior people have poor vision which makes it difficult for them to maker visual judgement and so calls for the use of Artificial Intelligence in visual image analysis and guided navigation systems.

In this project, we deal with two important enhancement features for mobility enhancement, Audio command and Vision aided obstacle detection and navigation.

We have implemented speech recognition algorithm using template of stored words for identifying the voice command given by the user. This frees the user of an agile hand to operate joystick or mouse control.

Also, we have developed a new appearance based obstacle detection system using stereo-vision cameras which estimates the distance of nearest obstacle to the wheel chair and takes necessary action. This helps user in making better judgement of route and navigate obstacles.

The main challenge in this project is how to navigate in an unknown/unfamiliar environment by avoiding obstacles.

# CONTENTS

# List of Figures

# CHAPTER 1

# INTRODUCTION

Longevity has been one of the greatest achievements of the 20th century in Health Care; as a result of which there are many more people benefitting from longevity today than ever before and their number is on the increase. On the other hand the rapid industrialization and urbanization has also brought about major changes in our social structure that adversely affects senior citizens. The centuries old joint family system has disintegrated and with it has collapsed the safety net of parents and grandparents. Consequently, more and more old people today, are either on their own or under the care of servants or worst left unattended. In order to address to their needs and be able to ameliorate some of the physical, mental and psychological problems, and also improve the quality of their lives, we need to equip them with certain technological innovations for assisted living in terms of better vision, hearing, mobility, navigation, available on easy command, which together enhance the mobility needed for them attending day to day chores more independently. Designing a versatile mobility enhancement system could be one such issue.

## 1.1 Motivation

Although powered wheelchairs provide a well-established solution for severely impaired persons, they do not cover all needs regarding mobility of people with even moderate impairment. For more artificial intelligence (AI) based judgement in navigation and man-machine interaction, we need to design a highly adaptable and modular mobility enhancement system to cover additional needs of the users.

The mobility functions need not only the conventional motorized chair, but also additional enhancements, that contribute to safe and guided mobility.

- Vision enhancement for easy surveying of surroundings and visitors

    This will be useful to people who have problems in neck or eyes.

- Friendly and personalized navigation aids by providing AI in judgement of visual information.

The features such as obstacle detection, stair case detection enable the user to avoid any accidents and have sensor aided navigation.

- Easy command system to chair

If the person has some disability with his hand, then audio command feature will be more helpful. It gives the user free movement of his hands and not ties them up in controlling the chair.

## 1.2 Objective

Keeping the above aspects in view, additional subsystems in the proposed enhancement device could either be a combination of digital camera, optoelectronic sensor for obstacle detection, geo-referencing for navigation, audio aids, microprocessors, suitable displays, An Information and Communication Technology (ICT) based equipment or a combination of these, with suitable mechanical add-ons interfaced for varied use and mass appeal. Overall, the enhanced mobility system should be

- ✓ Low cost & Ease of operation of control panel
- ✓ Navigation & obstacle recognition aids
- ✓ Audio-Vision aids

Out of these broad objectives in mind, we have implemented the speech command input part in the third semester as a minor project (that has already been evaluated and demonstrated) and the obstacle recognition aids in the final semester as the major project. Hence, in this thesis, the obstacle detection is discussed more in detail than the speech recognition part.

The main challenge in the obstacle detection is to tackle the following visual features: lighting conditions, stereo correspondence/disparity estimation, range, resolution, accuracy and speed of analysis. We have implemented our proposed algorithm in Matlab. Even though the Matlab is slower than C/C++, it proves the algorithm. The lighting conditions can vary from place to place. So, our algorithm should be robust enough to manage the change in lighting conditions and give same results all the time. The obstacle detection algorithm should perform well under various scenarios like change in location, change in time, change in environment conditions, changes in illumination, non-texture and transparent objects, eliminating the effect of shadows, etc. Also, the algorithm should be adaptive to those changes.

The stereo correspondence module is used to find the disparity value at each pixel which plays the major role in calculating the distance of the obstacle. There is no use of finding disparity at all pixels. Instead, we need disparity map in only the area concerned where the mobility of wheel chair is possible, thereby reducing time for estimating the distance. Also, the disparity estimation algorithm should give correct results under various conditions similar to the conditions for the obstacle detection algorithm.

The range, resolution and the accuracy of the distance estimation is essential as it is useful in predicting the obstacles in advance. The speed of the processing should be less and plays an important role as it is necessary to take action like changing the direction of wheelchair or stopping the chair well before hitting the obstacles, if any in the path. Hence altogether, fast and robust method should be used in all the sub modules of the project as the user fully trusts our algorithms for the movement of the wheelchair.

First, we discuss background works relating to Obstacle detection and then about speech recognition.

## 1.3 Related work in Obstacle Detection

While an extensive body of work exists for range-based obstacle detection, little work has been done in appearance based obstacle detection [4]. Interestingly, Shakey, the first autonomous mobile robot, used a simple form of appearance-based obstacle detection [5]. Because Shakey operated on textureless floor tiles, obstacles were easily detected by applying an edge detector to the monochrome input image.

However, Shakey's environment was artificial. Obstacles had non-specular surfaces and were uniformly coated with carefully selected colors. In addition, the lighting, walls, and floor were carefully set up to eliminate shadows. Horswill used a similar method for his mobile robots Polly and Frankie, which operated in a real time environment [6]. Polly's task was to give simple tours of the 7th floor of the MIT AI lab, which had a textureless carpeted floor.

Obstacles could thus also be detected by applying an edge detector to the monochrome input images, which were first subsampled to 64´48 pixels and then smoothed with a 3´3 low-pass filter.

Shakey and Polly's obstacle detection systems perform well as long as the background texture constraint is satisfied, i.e., the floor has no texture and the environment is uniformly illuminated. False positives arise if there are shiny floors, boundaries between carpets, or shadows. False negatives arise if there are weak boundaries between the floor and obstacles.

Turk and Marra developed an algorithm that uses color instead of edges to detect obstacles on roads with minimal texture [7]. Similar to a simple motion detector, their algorithm detects obstacles by subtracting two consecutive color images from each other. If the ground has substantial texture, this method suffers from similar problems as systems that are based on edge detection. In addition, this algorithm requires either the robot or the obstacles to be in motion. While the previously described systems fail if the ground is textured, optical flow systems actually require texture to work properly. A thorough overview of such systems is given by Lourakis and Orphanoudakis [8]. They themselves developed an elegant method that is based on the registration of the ground between consecutive views of the environment, which leaves objects extending from the ground unregistered. Subtracting the reference image from the warped one then determines protruding objects without explicitly recovering the 3D structure of the viewed scene.

However, the registration step of this method still requires the ground to be textured. In her master's thesis, Lorigo extended Horswill's work to domains with texture [9]. To accomplish this, her system uses color information in addition to edge information. The key assumption of Lorigo's algorithm is that there are no obstacles right in front of the robot. Thus, the ten bottom rows of the input image are used as a reference area. Obstacles are then detected in the rest of the image by comparing the histograms of small window areas to the reference area.

The use of the reference area makes the system very adaptive. However, this approach requires the reference area to always be free of obstacles. To minimize the risk of violating this constraint, the reference area cannot be deep. Unfortunately, a shallow reference area is not always sufficiently representative for pixels higher up in the image, which are observed at a different angle. The method performs in real-time, but uses an image resolution of only 64´64 pixels.

Similar to Lorigo's method, our obstacle detection algorithm also uses color information and can thus be used in a wide variety of environments. Like Lorigo's method, our system also uses histograms and a reference area ahead of the robot. In addition, our method provides binary obstacle images at high resolution in real-time.

In Stereo-Vision approach, a great deal of research has been done to study different algorithms of obtaining disparity map through which is an indication of distance in real world. Review of top ranked dense stereo algorithms and their corresponding classification is given in Scharstein and Szeliski's survey [3]. According to [3], Graph cut method outperforms all other stereo matching methods and provides excellent results in terms of disparity maps. However, Graph cut methods are quite computationally intense to calculate. Second best method in [3] that produces precise depth map is belief propagation. In [10] real time belief propagation stereo approach based on energy-minimisation optimisation is proposed. Even though results presented in this paper are more than acceptable, using only 16 levels of disparity is not enough for accuracy in estimating obstacle distance. Other Global minimisation based approaches are also possible. Alternatively, the Normalised Cross Correlation which is also discussed in [3] can be used as it is more simple to implement and relatively accurate too.

The Goal of this thesis is to implement obstacle detection using the algorithm in [2] with some modifications and distance estimation part using the Normalised Cross Correlation algorithm [11] to provide a robust approach.

## 1.4 Related work in Speech Processing

Coming to the Speech processing part [25], the transformation of speech into feature vectors is followed by the process of recognizing what was actually spoken. There are several approaches to this problem. These include: knowledge-based approaches (using neural networks), template matching (using Dynamic Programming DP/ Dynamic Time Warping DTW), stochastic approaches and connectionist approaches (using Hidden Markov Models HMM). These methods are not mutually exclusive.

The Speech recognition systems based on the DP algorithm using prototype vector Sequences to match against the unknown utterance (also known as dynamic pattern matching) were widely used in commercial products and also simple to understand the implementation algorithm. Hence in this project, we have implemented the Speech Recognition using Dynamic Pattern Matching approach as discussed in [25].

## 1.5 Organization of the report

The first Chapter explains the definition and the objective of the problem and the approach to the solution to the problem is presented.

Chapter 2 explains how we made the real-time setup of Obstacle detection and the working of the setup. First, we have to identify the obstacles in an unknown environment and then find the distance of nearest obstacle based on which we will control the wheel chair.

Chapter 3 presents the algorithm of detecting the obstacles using a single color camera in an easy and fast manner.

In Chapter 4, the basics of the stereo-vision and how we are finding the distance of the obstacle using stereo vision is described in detail.

Chapter 5 deals with the complete Speech processing part of the project with the theories behind voice recognition and the approach.

The chapter 6 shows the various results of the project, accuracy and real time ability of the proposed method is discussed.

Finally, we conclude the success of the project in Chapter 6 with open topics for future scope.

# CHAPTER 2

# SYSTEM SETUP FOR OBSTACLE DETECTION

The first section in this chapter explains about the model of the camera used and the calibration procedure needed for stereo vision purpose. The next section deals with the software GUI part which helps us to visualize the results at various stages. The final section presents the circuit used for interfacing the computer and the robot.

We made an experimental setup with two cameras placed near to each other in a small robot arrangement which can move front/back/right/left. The analysis of images is done using Matlab software. Based on the obstacle distance, necessary action such as, moving the robot forward or stopping or turning left/right will be taken. The robot motors are interfaced with the PC through a microcontroller and motor drivers. Fig 2.1 shows the robot prototype developed. Motor1 is used for turning the robot left or right and Motor2 used for moving the robot forward or backward. Two cameras (Left and Right) are used for stereo vision.

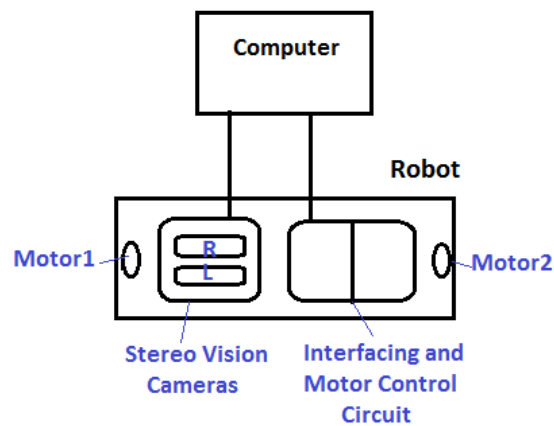

Figure 2.1 System Setup

## 2.1 Camera model and Calibration

For the stereo system, we used two Microsoft H5D-003 Lifecam Cinema web cameras which have high definition 720p resolution and baseline distance is 50mm.

The camera is fixed well such that the repetition in calibration procedure can be avoided. The exposure setting for the camera is chosen for optimized lightning condition of the environment. During initial phase of the project such as calibration and algorithm checking, the cameras are mounted on a camera folder and fixed at a particular position. This arrangement is shown in the figure 2.2.

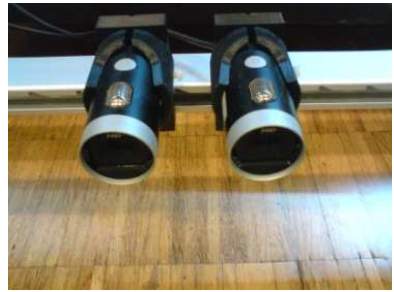

Figure 2.2: Camera Setup

Camera calibration is very important for validating real time measurements. Calibration refers to the relation between camera's natural units (pixels) and the physical world (meters) and is a critical component in finding the depth of a point.

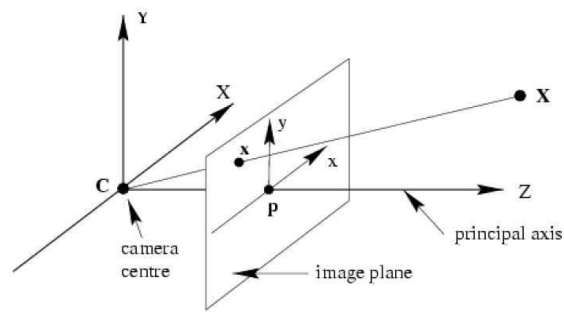

Figure 2.3: Pinhole Camera model [1]

### 2.1.1 Camera model

The simplest dorm of real camera consists of a pinhole and an image plane as shown in figure 2.3. A pinhole camera model assumes that all projection rays from the camera intersect at a single point known as the camera centre. The relation between real world coordinates P(X,Y,Z) and the camera image plane p(x,y) in pinhole camera is

$$x = fX/Z$$
$$y = fY/Z$$

(2.1)

Where f is the focal distance of the lens. Using homogeneous coordinates, P and p can be represented by $(X,Y,Z,1)^T$ and $(fX/Z, fY/Z, 1)^T$. Represented as homogeneous vectors, the mapping from 3D to 2D space can be expressed in matrix form as,

$$p = \begin{bmatrix} f & 0 & 0 & 0 \\ 0 & f & 0 & 0 \\ 0 & 0 & 1 & 0 \end{bmatrix} \begin{bmatrix} X \\ Y \\ Z \\ 1 \end{bmatrix}$$

(2.2)

The principal point is the intersection of the camera's optical axis with the image plane. The above formulation assumes that the origin of coordinates in the image plane coincides with the principal point. Assuming that the coordinate of the principal point is $(x_0, y_0)^T$ we can write

$$p = \begin{bmatrix} f & 0 & x_0 & 0 \\ 0 & f & y_0 & 0 \\ 0 & 0 & 1 & 0 \end{bmatrix} \begin{bmatrix} X \\ Y \\ Z \\ 1 \end{bmatrix}$$

(2.3)

Pinhole camera is characterized by two sets of parameters.

**Intrinsic parameters** – describes the internal geometry and the optical characteristics of the camera. We introduce a matrix Called intrinsic matrix of the camera.

$$A = \begin{bmatrix} f & 0 & x_0 \\ 0 & f & y_0 \\ 0 & 0 & 1 \end{bmatrix}$$

(2.4)

**Extrinsic parameters** – describe the camera position and the orientation on the real world. The world coordinate system is transformed into the camera coordinate system through a rotation R and a translation t. (R,t) are called the extrinsic parameters. The relationship between a world point P(X,Y,Z) and its projection p(x,y) is given by

$$p = A\,[Rt]\,P$$

(2.5)

To compute a comparison between two images captured from two different cameras, intrinsic and extrinsic parameters are fundamental. The pinhole model is an ideal camera model. It does not take into consideration distortion effects introduced by real lenses. The major components of the lens distortion are radial and tangential distortion.

Four parameters are used to describe the distortion: k1 and k2 are radial distortion coefficients: p1 and p2 are tangential distortion coefficients. After the calibration process of a camera, the distortion parameters are known and can be used to correct the distortion.

### 2.1.2 Camera Calibration

We have presented a camera model describing the projection of the real world coordinates co-ordinates into camera image. Now, parameters describing this camera model have to be determined. This is called intrinsic camera calibration [24].

The previous section introduced four internal parameters to describe a camera: f, the focal length in pixel units. The coordinates in the camera image of the optical center $x_0, y_0$. These parameters are completed by four distortion parameters: k1,k2 for radial distortion and p1,p2 for tangential distortion.

Moreover six extrinsic parameters describe the position of the camera in the real world coordinate system: three parameters for the rotation R and three for translation t. Some of the intrinsic parameters can be retrieved from the camera specification sheet, but due to manufacturing mechanics they are very inaccurate. Thus, these parameters must be determined for each camera with precision. This process is called calibration. The main parameters to be determined are

- f, focal length of the camera
- t, Translational vector which gives separation between two cameras in X, Y and Z axis.

Only these two parameters are used in calculating actual distance of a point in the image. Other parameters such as Distortion and rotational vector are computed to make sure that the stereo setup is proper.

We use the OpenCV library to calibrate the process. In this library, the algorithm of Zhang is used [13]. This algorithm requires the camera to observe a planar pattern at different orientations. The main idea of Zhang's algorithm is to estimate a homography between the model plane and its image in the camera for each view. A homography is a mathematical relation between two figures, so that any given point in one figure corresponds to one and only one point in the other and vice versa. Feature points of the planar pattern are detected in the images and associated to feature points of the model plane, using a technique based on maximum likelihood criterion, a homography is estimated for each observation, mapping the model image to the camera image. The extraction of the feature points can be easily automated in some cases. The OpenCV library provides a function to extract the corners of a chessboard pattern. Figure 2.4 shows an image of the chessboard pattern for calibration and the corners detected in that image [14].

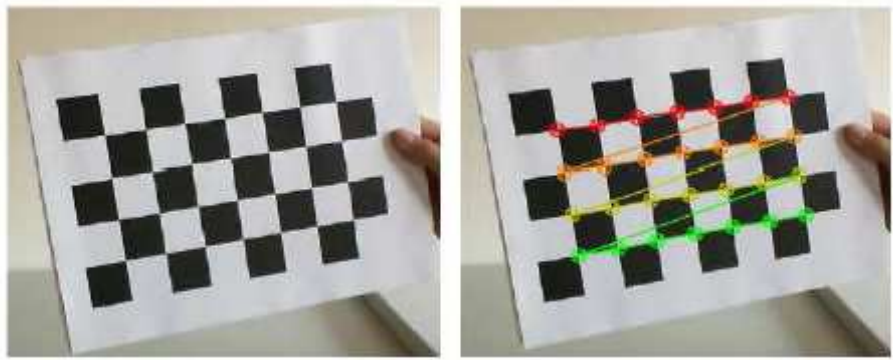

Figure 2.4: Calibration using Chessboard pattern

Zhang's algorithm first starts with an analytical solution. This analytical solution is computed using the linear part of the camera model (without distortion). This solution is then optimized suing a nonlinear solution; the distortion parameters are finally estimated [14].

## 2.2 Matlab GUI

A Graphical User Interface (GUI) is developed using Matlab for display of the complete results. The snapshot of the GUI is shown in figure 2.5. It has four image preview windows. Two images are for live video preview from two cameras called

Left and Right cameras. The third image for showing the result of Obstacle detection (Binary image) in which the white spot represents obstacles and the black spot represents ground region (non-obstacles). We use left camera image for finding Binary segmented image. The fourth image window is for displaying the disparity map in which, brighter the point in the image is, closer the actual distance of the point in real world. The Figure 2.5 shows the snapshot of the GUI.

The concepts behind how obstacle is detected and computing disparity map are described in the next two chapters. The disparity map is calculated in only particular region where the robot can move assuming it to be moving in forward direction. This region is shown in the Figure 2.6. We calculate disparity using both the left and right camera images, for which simultaneous image snapshot is taken from both the cameras.

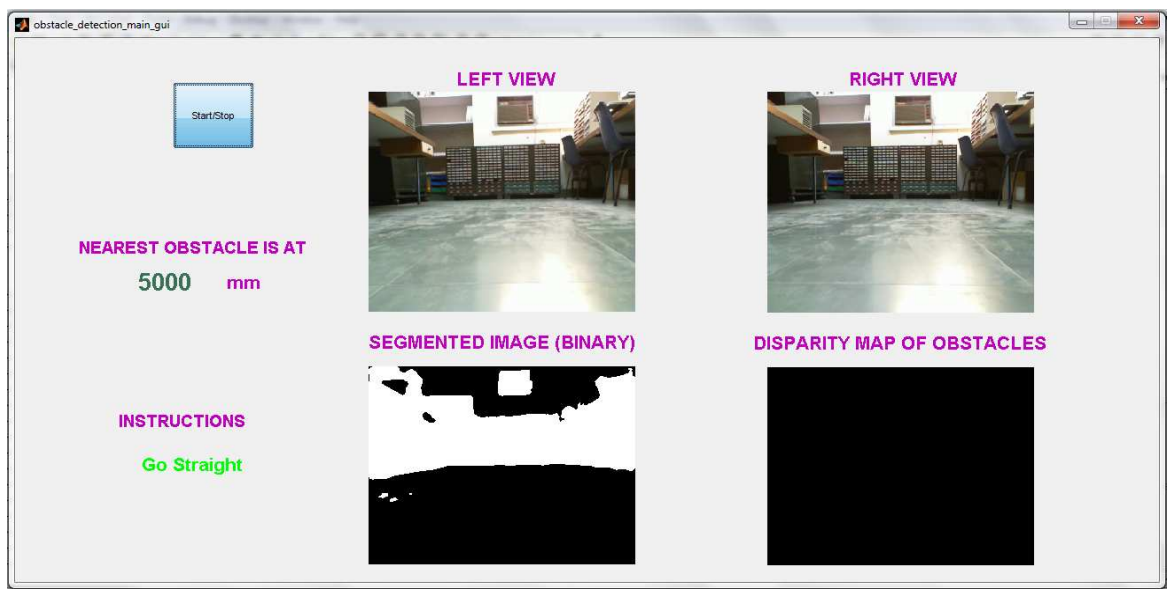

Figure 2.5: Matlab GUI

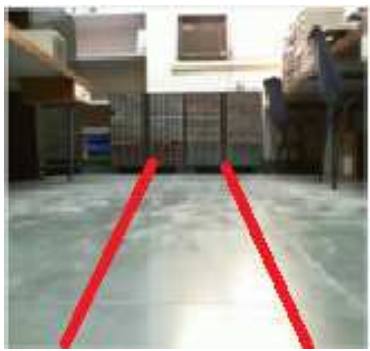

Figure 2.6: Region of navigation

The distance of the nearest obstacle (D) is found and displayed. Based on that distance value, control instructions will be sent to robot, based on which the robot may move forward, stop, turn right or left. The algorithm used to control the robot is given below.

If D>750mm, then Go Straight (No problem with obstacles)

If 600mm<D<=750mm, then turn right or left (To avoid obstacle)

If D<=600mm, then Stop (because obstacle is very close)

The distance of near obstacles in both right and left sides of the image is averaged and compared. If right side is having more close obstacles then left side, then Turn slightly left instruction will be sent to the robot and vice versa.

## 2.3 Test setup

As shown in the figure 2.1, the robot prototype developed has two motors, one for moving forward or reverse and another for turning left or right direction. These two motors are controlled by Half H bridge driver IC (L293D). The motor control signal is given from a PIC microcontroller (16F877A) which acts as interface between the computer and the robot. The microcontroller is interfaced with serial port of the PC through serial to CMOS level converter IC (Max232). The schematic circuit for interfacing is shown in figure 2.7.

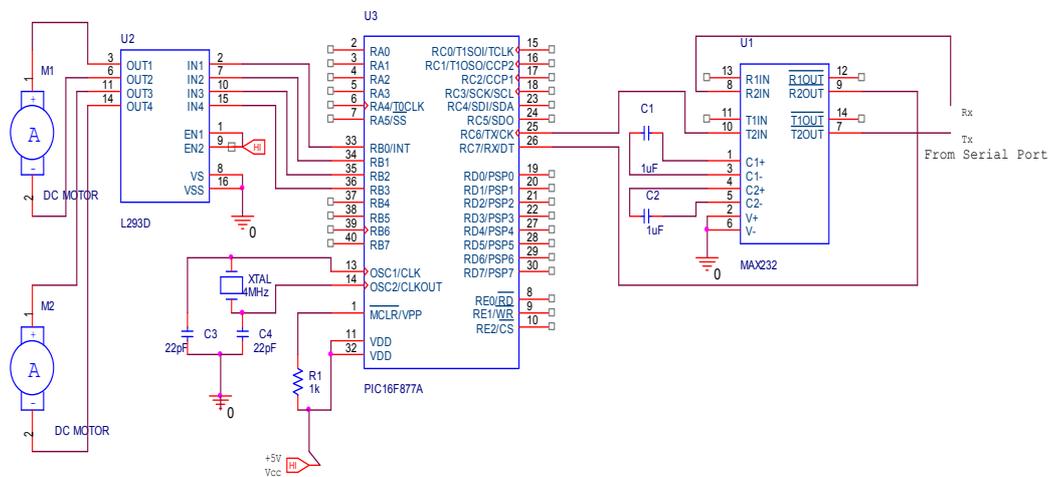

Figure 2.7: Interfacing Circuit

The purpose of the microcontroller is to get the command from the computer and give necessary output to the motor driver. The IC details are given in the appendix along with the program written in the microcontroller.

The figure 2.8 shows the prototype developed for the obstacle detection. The cameras are mounted on a robot base having two motors. The PCB for interfacing circuit is also embedded in the prototype.

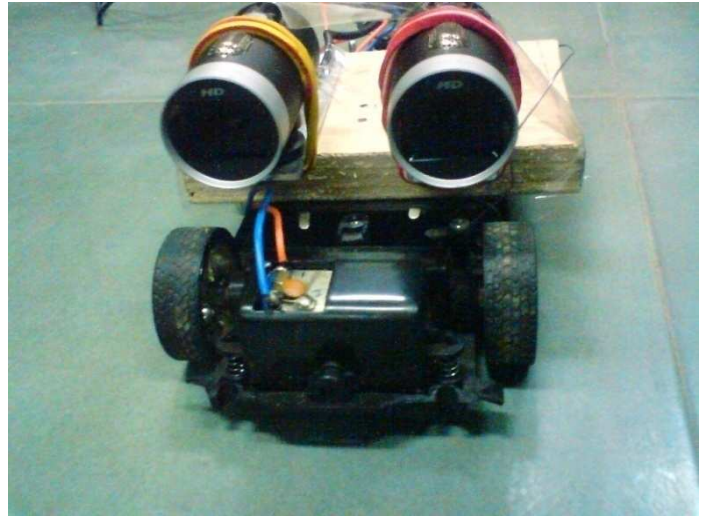

Figure 2.8: Prototype developed

We have discussed about the setup of the obstacle detection system dealing with the camera model used, the calibration procedure, software interface and the hardware circuit for the interfacing of the developed prototype with the computer.

The next chapter describes how we are finding the position of the obstacles. After detecting the position of the obstacles, the calibration parameters are useful in finding distance of the obstacle which is discussed in the chapter 4.

# CHAPTER 3

# OBSTACLE DETECTION

This chapter presents a new vision-based obstacle detection method for mobile robots based on the work of Iwan Ulrich and Illah Nourbakhsh [2].

The basic principle we use for obstacle detection is that, in an image, each individual pixel is classified as belonging either to an obstacle or the ground based on its colour appearance. The method uses a single passive color camera, performs in real-time, and provides a binary obstacle image at high resolution. The difference between our proposed algorithm and the paper by Ulrich [2] is that we will estimate the distance of obstacle using stereo-vision for removing certain assumptions which is explained at a later stage.

## 3.1 Background

Obstacle detection is an important task for many mobile robot applications. There are two types of obstacle detection.

- Range based Obstacle detection using range sensors
- Appearance based obstacle detection sing cameras

Most mobile robots rely on range data for obstacle detection. Popular sensors for *range-based* obstacle detection systems include ultrasonic sensors, laser rangefinders, radar, optical flow, and depth from focus. Because these sensors measure the distances from obstacles to the robot, they are inherently suited for the tasks of obstacle detection and obstacle avoidance. However, none of these sensors is perfect. Ultrasonic sensors are cheap but suffer from specular reflections and usually from poor angular resolution. Laser rangefinders and radar provide better resolution but are more complex and more expensive. Most depth from X vision systems requires a textured environment to perform properly. Moreover, optical flow is computationally expensive.

In addition to their individual shortcomings, all range based obstacle detection systems have difficulty detecting small or flat objects on the ground. Reliable detection of these objects requires high measurement accuracy and thus precise calibration. Range sensors are also unable to distinguish between different types of ground surfaces. This is a problem especially outdoors, where range sensors are usually unable to differentiate between the sidewalk pavement and adjacent flat grassy areas. While small objects and different types of ground are difficult to detect with range sensors, they can in many cases be easily detected with color vision. For this reason, we have developed a new *appearance-based* obstacle detection system that is based on passive monocular color vision. The heart of our algorithm consists of detecting pixels different in appearance than the ground and classifying them as obstacles.

The algorithm performs in real-time, provides a high-resolution obstacle image, and operates in a variety of environments. The algorithm is also very easy to train. The fundamental difference between range-based and appearance-based obstacle detection systems is the obstacle criterion. In range-based systems, obstacles are objects that protrude a minimum distance from the ground. In appearance-based systems, obstacles are objects that differ in appearance from the ground.

## 3.2 Appearance-Based Obstacle Detection

Our obstacle detection system is purely based on the appearance of individual pixels. Any pixel that differs in appearance from the ground is classified as an obstacle. The three assumptions used in [2] are

1. Obstacles differ in appearance from the ground.
2. The ground is relatively flat.
3. There are no overhanging obstacles.

The first assumption allows us to distinguish obstacles from the ground, while the second and third assumptions allow us to estimate the distances between detected obstacles and the camera. But, the second and third assumption is not always suitable for working in both indoor and outdoor environments.

Therefore, in our method, we use stereo vision algorithm for estimating the distance of the obstacle so that we remove assumption 2 and 3 and make the approach more robust. The classification of a pixel as representing an obstacle or the ground can be based on a number of local visual attributes, such as intensity, color, edges, and texture. It is important that the selected attributes provide information that is rich enough so that the system performs reliably in a variety of environments. The selected attributes should also require little computation time so that real-time performance can be achieved without dedicated hardware. The less computationally expensive the attribute, the higher the obstacle detection update rate, and the faster a mobile robot can travel safely.

To best satisfy these requirements, we decided to use color information as our primary cue. Color has many appealing attributes, although little work has lately been done in color vision for mobile robots. Color provides more information than intensity alone. Compared to texture, color is a more local attribute and can thus be calculated much faster. Systems that solely rely on edge information can only be used in environments with textureless floors, as in the environments of Shaky and Polly. Such systems also have more difficulty differentiating between shadows and obstacles than color based systems. For many applications, it is important to estimate the distance from the camera to a pixel that is classified as an obstacle.

### 3.3. Basic Approach

Our appearance-based obstacle detection method consists of the following four steps:

1. Filter color input image.
2. Transformation into HSI color space.
3. Histogramming of reference area.
4. Comparison with reference histograms.

In the first step, the 320´240 color input image is filtered with a 5´5 Gaussian filter to reduce the level of noise. In the second step, the filtered RGB values are transformed into the HSI (hue, saturation, and intensity) color space.

Because color information is very noisy at low intensity, we only assign valid values to hue and saturation if the corresponding intensity is above a minimum value. Similarly, because hue is meaningless at low saturation, hue is only assigned a valid value if the corresponding saturation is above another minimum value. An appealing attribute of the HSI model is that it separates the color information into intensity and a color component. As a result, the hue and saturation bands are less sensitive to illumination changes than the intensity band.

In the third step, a trapezoidal area in front of the mobile robot is used for reference. The valid hue and intensity values of the pixels inside the trapezoidal reference area are histogrammed into two one-dimensional histograms, one for hue and one for intensity. The two histograms are then low-pass filtered with a simple average filter.

Histograms are well suited for this application, as they naturally represent multi-modal distributions. In addition, histograms require very little memory and can be computed in little time. In the fourth step, all pixels of the filtered input image are compared to the hue and the intensity histograms. A pixel is classified as an obstacle if either of the two following conditions is satisfied:

i) The hue histogram bin value at the pixel's hue value is below the hue threshold.

ii) The intensity histogram bin value at the pixel's intensity value is below the intensity threshold.

If none of these conditions are true, then the pixel is classified as belonging to the ground. In the current implementation, the hue and the intensity thresholds are not set to particular number of pixels. Rather, they are set to values which is one $50^{th}$ of the maximum value in that histogram. The number of Histogram bins used is 5. These values have been adapted after repeated trial experiments.

## 3.4 Advantages of Appearance based Obstacle detection

- Textures in floors is not a problem
- Less sensitive to Shadows & illumination changes/reflections
- Less computation time & memory
- Simple to implement in software and hardware
- Easy to implement adaptive algorithm with neural networks
- Useful for further analysis like finding the right path to navigate

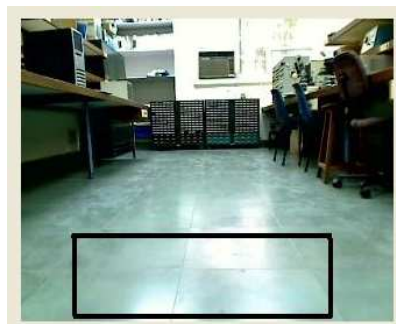

Figure 3.1 Reference area

Our algorithm performs quite well in different environment. The reference area considered for analysis is shown in figure 3.1. Independent of the lightning condition the algorithm is able to detect obstacles. The algorithm also rejects the effect of shadows. These qualities can be observed in Figure 3.2.

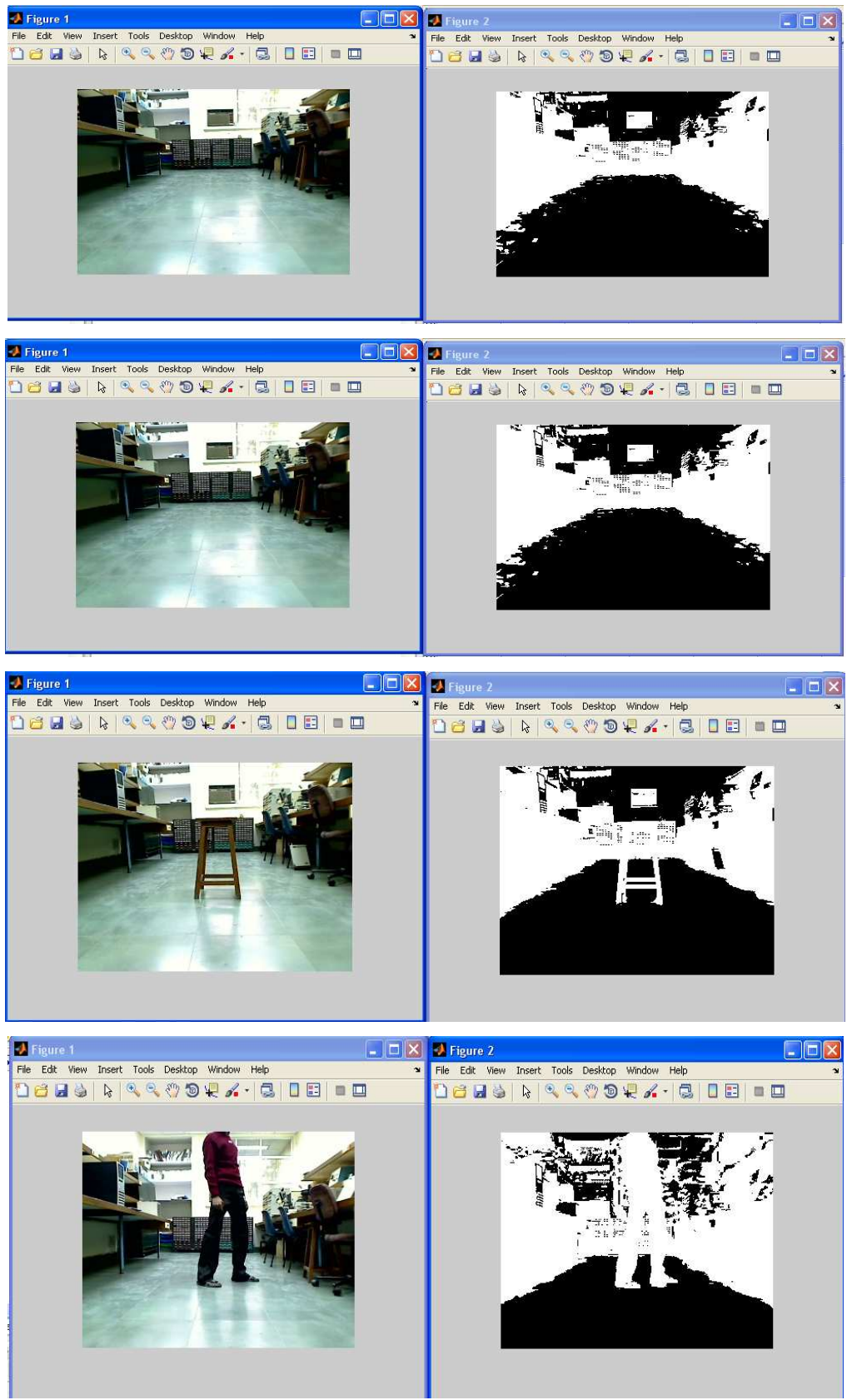

Figure 3.2: Obstacle detection in various environment conditions

In particular, the algorithm also detects the cable lying on the floor, which is very difficult to detect with a range-based sensor. Also, it is able to detect the narrow paths to navigate. These points can be obvious in Figure 3.3.

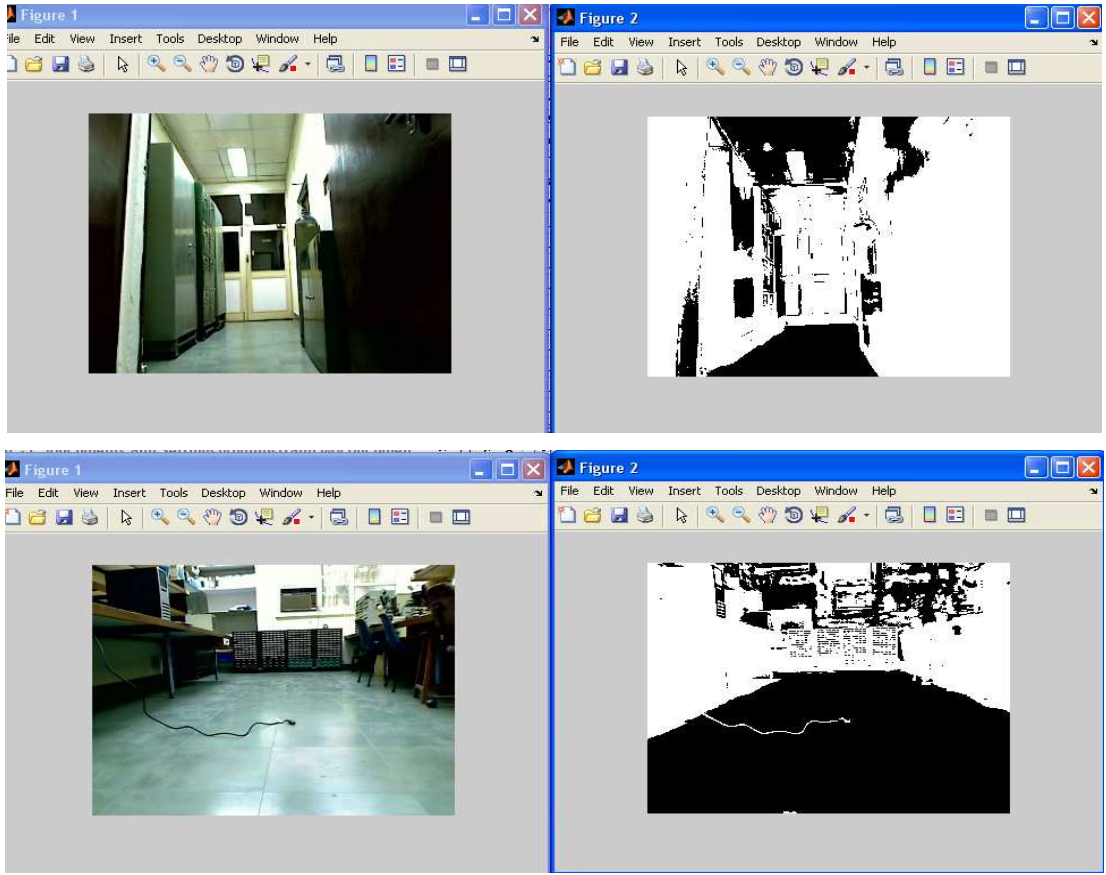

Figure 3.3: Detecting Small/narrow obstacles

Thus we have seen how we can find detect the obstacles using appearance based obstacle detection principle. The algorithm gives a binary image in which bright region represents obstacles and dark region is ground. In the next chapter, we will discuss how to find the actual distance of the point where nearest obstacle is present.

# CHAPTER 4

# DISTANCE ESTIMATION

This chapter presents the brief introduction, principle and concepts of the stereo vision, disparity map estimation and distance calculation from disparity value. We use the stereo vision/triangulation principle to calculate the distance of the obstacles.

## 4.1 Stereo Vision

A binocular stereo vision system is composed of two cameras both of which observe the same scene. The main task of stereo vision is to compute three dimensional data from these 2D input images. Computer stereo vision tries to imitate the human visual system. The human visual system obtains information about the world through two planar images that are captured on the retina of each eye. The position of a scene point in right view is horizontally shifted in the left view. This displacement commonly referred to as disparity, human brain use to deduce the depth information of the scene. Although this course of action appears simple, for computers is surprisingly difficult.

The major challenge that one faces in computer vision is solving the correspondence problem. It describes the risk of automatically computing the correct disparity at each pixel.

## 4.2 Stereo Triangulation

The same triangulation method that is used in navigation and surveying is used to calculate depth. Basic triangulation uses the known distance of two separated points looking at the same scene point. From these parameters the distance to the scene point can be calculated.

This same basic idea is used in stereo vision to find depth information from two images. Figure 4.1 graphically shows the geometry.

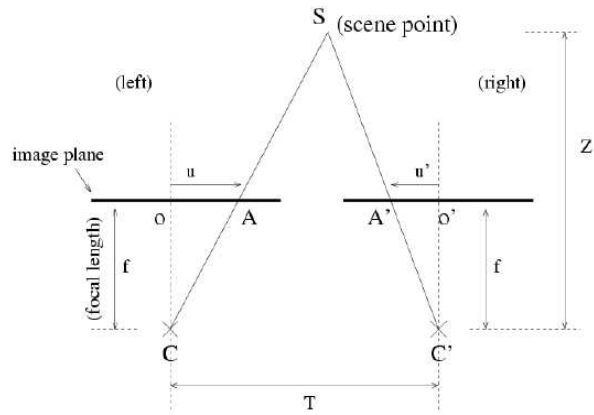

Figure 4.1: Triangulation in a Stereo Vision system

In the above arrangement, two cameras (*C,C'*) see the same feature point (*S*). The location of the point in the two image planes is denoted by *A* and *A'*. When the cameras are separated by a distance T, the location of A and A' from the cameras normal axis will differ (denoted by *U,U'*). Using these differences, the distance (*Z*) to the point can be calculated from the following formula:

$$Z = f \frac{T}{U - U'}$$

(4.1)

In order to calculate depth however, the difference of *U* and *U'* need to be established. The image analysis techniques used to find the differences (U-U') in the images are the focus of the next section.

We know the focal length of the camera (f) and camera baseline distance (T) using the calibration procedure explained by Zhang [13]. Thus, if we know disparity value, then the actual distance of the real world point is known.

## 4.3 Epipolar Geometry and Rectification

### 4.3.1 Epipolar Constraint

We get two images from two different angles of view. It can be verified that for any 3D point, its projection in one image must lie in the plane formed by its projection in the other image and the optical centres of the two cameras. This is known as Epipolar constraint.

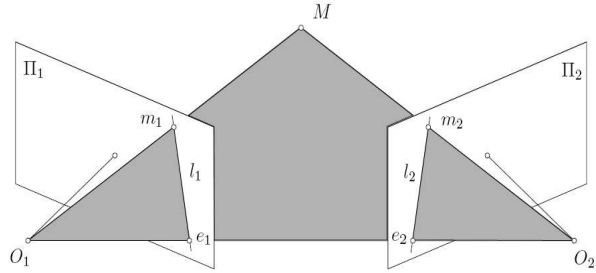

Figure 4.2: Epipolar geometry

Assuming that a 3D point M is observed by two cameras with optical centres O1 and O2, we get the projections m1 and m2 of this point in the two image planes. We define epipolar plane as the plane containing M, O1 and O2. Note that m1, m2 also belong to this plane. Consider the case where m2, O1 and O2 are given and we want to find the corresponding point of m2 in the first image, i.e.: m1. The epipole plane is determined by m2, O1 and O2 (without knowing the position of M). Since m1 must belong to this epipole plane and the image plane of the first camera, it must lie on the line l1 which is the intersection of these two planes.

We call the line l1 the epipolar line associated with m2. By symmetry, the m2 must lie on the epipolar line l2 associated with m1. This epipolar constraint is used to look for the corresponding point in one image given a point in the other image. The search can be restricted to the epipolar line instead of the whole image i.e. the correspondence search can be reduced to 1D search task [12].

**4.3.2 Epipolar Rectification**

Especially interesting case arises if both image planes L and P lie in a common plane and their x-axes are parallel to the baseline, which is shown in Figure 4.3. In this configuration, the epipoles move to infinity and the Epipolar lines coincide with horizontal scanlines. The matching point of a pixel in one view can then be found on the same scanline in the other view, i.e: yl=yr where yl and yr refer to the y-coordinated of a scene point in the left and right images respectively. The horizontal offset between corresponding pixels xl-xr is referred to as disparity [1]. To take benefit of this simple geometry, the images of two cameras in general positions can be reprojected onto a plane that is parallel to the baseline. This process is known as rectification or Epipolar rectification. The rectification step involves resampling of the image, and therefore some precision in the 3D construction is lost.

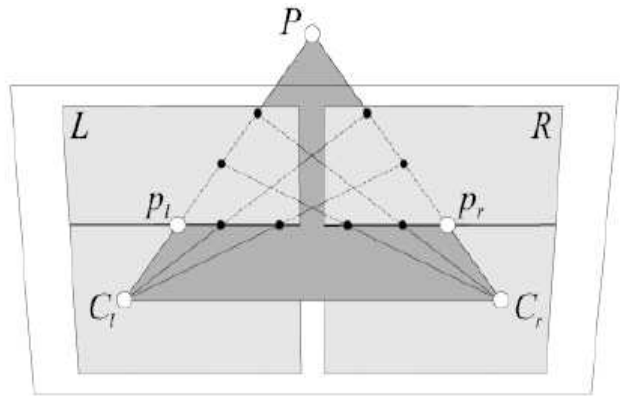

Figure 4.3: Epipolar lines after epipolar rectification

However, since it is more convenient to search for correspondences along horizontal scanlines than to trace general Epipolar lines, this transformation is commonly applied. In fact, Most of stereo matching algorithms work with rectified images [12]. In our project, we assume that the images are epipolar by having proper mechanical setup so that we don't have to rectify the images explicitly.

## 4.4 Stereo Correspondence

In order to compute accurate correspondence between two images, it is necessary to identify possible issues that make this task more difficult and to establish right assumptions relating to those issues.

The first issue while preparing proper setup for recording stereo images is color/intensity change between pixels originating from same 3D point of the scene. Establishing the same lighting conditions for two viewpoints turn out to be quite challenging. Wide baseline between two cameras introduces different intensity values of corresponding pixels which can produce false matches. Since we want to avoid this problem, we have to achieve same illumination of two views. This can be done by modelling diffused lighting. In our application, most of the indoor/outdoor lighting will be diffused one and so there is less lighting problem unless we have a power cut.

The second problem is that, it is possible to have uncertain matching in an untextured area due to many possible corresponding points of almost same intensity model. Matching pairs of some points from left view cannot be uniquely found in right view. This problem can be solved to some extent if we use block matching approach instead of pixel to pixel matching approach.

Finally, we have to consider the case when some points are visible only from one view. We define such pixels as occluded pixels. This problem is depicted on figure 4.4. Since occlusions often occur only at depth discontinuities, it won't be a problem in our application.

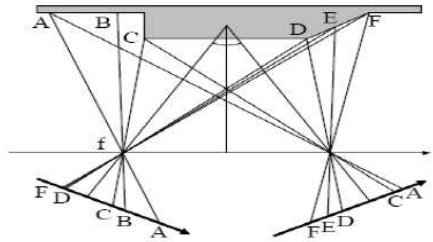

Figure 4.4: Occluded Regions [15]

**4.4.1 Matching constraints**

In order to minimum false matches, some matching constraints must be imposed. Below is a list of the commonly used constraints.

- **Similarity** [16]: For the intensity-based approach, the matching pixels must have similar intensity values (i.e. differ lower than a specified threshold) or the matching windows must be highly correlated. For the feature-based approach, the matching features must have similar attribute values.
- **Uniqueness** [17]: Almost always, a given pixel or feature from one image can match *no more than one* pixel or feature from the other image.
- **Continuity** [17]: The cohesiveness of matters suggests that the disparity of the matches should vary smoothly almost everywhere over the image. This constraint fails at discontinuities of depth, for depth discontinuities cause an abrupt change in disparity. Smoothness is assumed by almost every correspondence algorithm either in implicit or explicit way.
- **Ordering** [18]: If m <-> m' and n <-> n' and if *m* is *to the left* of *n* then *m'* should also be *to the left* of *n'* and vice versa. That is, the ordering of features is preserved across images. The ordering constraint fails at regions known as *the forbidden zone (figure 4.5).*

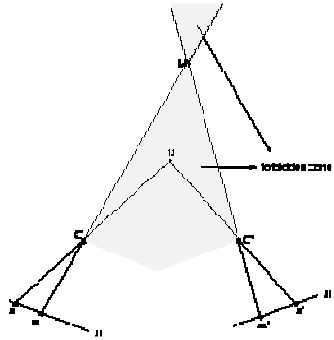

Figure 4.5: Ordering constraint and Forbidden zone

- **Epipolar:** Given a feature point *m* in the left image, the corresponding feature point *m*' must lie on the corresponding epipolar line. This constraint reduces the search space from two-dimensions to one-dimension. Unlike all the other constraints, the epipolar constraint would never fail and could be applied reliably once the epipolar geometry is known.
- **Relaxation**. A global matching constraint to eliminate false matches.

## 4.5 Disparity

As mentioned above, differences between two images gives depth information. These differences are known as disparities. The key step to obtaining accurate depth information is therefore finding a detailed and accurate disparity map. Disparity maps can be visualised in greyscale. Close objects result in a large disparity value. This is translated into light greyscale values. Objects further away will appear darker.

Obtaining depth information is achieved through a process of four steps. Firstly the cameras need to be calibrated. After calibrating the cameras the assumption is made that the differences in the images are on the same horizontal or *Epipolar* line [19].

The secondly step is the decision as to which method is going to be used to find the differences between the two images. Once this decision is made, an algorithm to obtain the disparity map needs to be designed or decided on. The third step is to implement the algorithm to obtain the disparity information.

The final step is to use the disparity information, along with the camera calibration set in step one, to obtain a detailed three dimensional view of the world.

This report focuses on the basic ideas behind the algorithms used to obtain disparity information. The other steps are relatively straight forward in their operation and implementation. It should be noted that even within the algorithms described below; there is ongoing research and therefore many different implementations.

There are many algorithms used to find the disparity between the left and right images. Additionally, there is a large amount of ongoing research into finding quicker and more accurate algorithms. However there are two commonly used algorithms that are currently used to find disparity. The first method is *feature-based* [20]. The second method is an *area-based* statistical method. Because they are widely used, we will focus on these two methods in this report.

**4.5.1 Feature Based Disparity**

This method looks at features in one image and tries to find the corresponding feature in the other. The features can be edges, lines, circles and curves. Nasrabadi [21] applies a curve segment based matching algorithm. Curve segments are used as the building block in the matching process.

Curve segments are extracted from the edge points detected. The centre of each extracted curve is used as the feature in the matching process. Medioni and Nevatia [22] uses segments of connected edge points as matching primitives. Stereo correspondence is achieved through minimising the differential disparity measure for global matching, by taking into account things such as end points and segment orientation.

For each feature in the left hand image ($Q_L$) there needs to be a similar feature in the right hand image ($Q_R$). A measure of similarity is needed to associate the two features. This measure is given by the following formula adapted from Candocia and Adjouadi [20].

$$\Psi(L \rightarrow R) = \frac{1}{N_L} \sum_{q=1}^{h} \frac{1}{(D_q + 1)}$$

(4.2)

Where $N_L$ = total number of features in the left image. $1/N_L$ is the weight associated with a matched feature. *h* is the minimum number of features found in either image, ie *h* = min($N_L, N_R$). $D_q$ is the minimum distance between a matched feature in the left and right images.

The features in the right image, within a constrained search area, with the highest similarity coefficient over a threshold are associated. These are then compared globally to other associated features to check for consistency. The difference in location of these features gives the disparity.

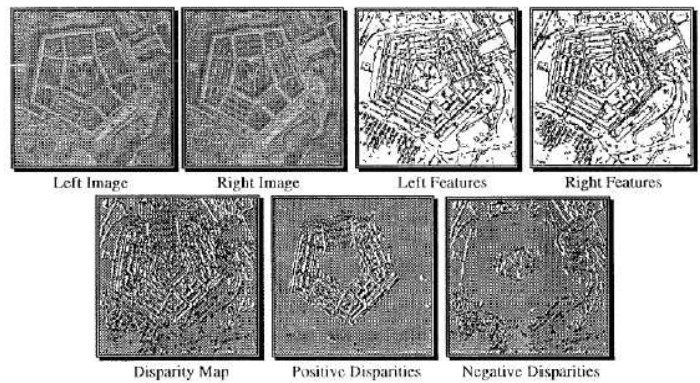

Figure 4.6: Results of the above feature based algorithm [20]

### 4.5.2 Area Based Disparity

This method is also called as Block matching. There are various techniques that are used in this algorithm. In both these methods a window is placed on one image. The other image is scanned using the same size window. The pixels in each window are compared and operated on. These are then summed to give a coefficient for the centre pixel. These techniques have been developed by Okutomi and Kanade [23].

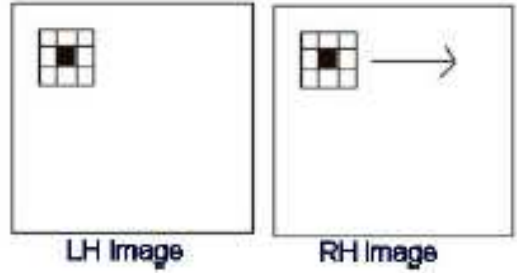

Figure 4.7: Area based disparity estimation

The first operation described is correlation. The output of the scanning window is convolved with the first and the location that gives the highest convolution coefficient is deemed to be the corresponding area. The correlation coefficient is given by:

$$C_{LR} = \sum_{[i,j] \in Window} L(i,j) R(i,j)$$

(4.3)

The second method uses the same window principle, but uses the Sum of Absolute differences (SAD) or Sum of squared differences (*SSD*) or Sum of Hamming Distances (SHD) or Normalised Cross Correlation (NCC), etc. This examines the pixel values in both windows and estimates the disparity by calculating the corresponding (SAD/SSD/SHD/NCC) coefficients. In this method the value of the coefficient needs to be minimised. The formula for SAD and NCC methods are given below.

$$SAD(x,y) = \sum_{i=1}^{M} \sum_{j=1}^{N} |T(i,j) - C(x+i, y+j)|$$

(4.4)

$$NCC(x,y) = \frac{\sum_{i=1}^{M} \sum_{j=1}^{N} I(x+i, y+j) \cdot T(i,j)}{\sqrt{\sum_{i=1}^{M} \sum_{j=1}^{N} I(x+i, y+j)^2} \cdot \sqrt{\sum_{i=1}^{M} \sum_{j=1}^{N} T(i,j)^2}}$$

(4.5)

Where, I or C is Source image (Left view). T is the Template image (Right view).

We have observed that the NCC method gives better results than other methods because NCC method is based on similarity between two image patches, while the *SAD* and *SSD* provide the measures for the total differences between two images. Also, NCC algorithm is more robust than SAD and SSD under linear illumination changes, so the NCC measure has been widely used in object recognition and industrial inspection [11]. Hence, in this project, we implemented the NCC block matching algorithm (eq.4.5) and obtained better results for disparity map compared to SAD/SSD.

Thus, the NCC algorithm gives the disparity value (U-U'). The focal length f and baseline T are known from calibration. Hence, by using eq. 4.1, we can find the distance of the obstacle. In the Chapter 5, we can see that the f and T are found from the calibration and the disparity map is displayed in the GUI. Using these values, the distance of obstacle is calculated and the nearest obstacle distance is displayed.

# CHAPTER 5

# SPEECH RECOGNITION

The correct recognition of words spoken by comparing with a library of stored words is called voice recognition. The process used to implement voice recognition is called speech processing. In this chapter, we will discuss the basics behind the speech processing algorithm [25] and how we have implemented the module.

## 5.1 The Speech Signal

### 5.1.1 Production of Speech

While you are producing speech sounds, the airflow from your lungs first passes the glottis and then your throat and mouth. Depending on which speech sound you articulate, the speech signal can be excited in three possible ways:

• **Voiced excitation:** The glottis is closed. The air pressure forces the glottis to open and close periodically thus generating a periodic pulse train (triangle–shaped). This fundamental frequency usually lies in the range from 80Hz to 350Hz.

• **Unvoiced excitation**: The glottis is open and the air passes a narrow passage in the throat or mouth. This results in a turbulence, which generates a noise signal. The spectral shape of the noise is determined by the location of the narrowness.

• **Transient excitation**: A closure in the throat or mouth will raise the air pressure. By suddenly opening the closure the air pressure drops down immediately. ("Plosive burst"

With some speech sounds these three kinds of excitation occur in combination. The spectral shape of the speech signal is determined by the shape of the vocal tract (the pipe formed by your throat, tongue, teeth and lips). By changing the shape of the pipe (and in addition opening and closing the air flow through your nose) you change the spectral shape of the speech signal, thus articulating different speech sounds.

### 5.1.2 Technical Characteristics of the Speech Signal

An engineer looking at (or listening to) a speech signal might characterize it as follows,

- ✓ The bandwidth of the signal is 4 kHz.
- ✓ The signal is periodic with a fundamental frequency between 80 Hz and 350 Hz.
- ✓ There are peaks in the spectral distribution of energy at

$$(2n - 1) * 500 \text{ Hz}; n = 1, 2, 3, \ldots \tag{5.1}$$

- ✓ The envelope of the power spectrum of the signal shows a decrease with increasing frequency (-6dB per octave)

#### 5.1.2.1 Bandwidth

The bandwidth of the speech signal is much higher than the 4 kHz stated above. In fact, for the fricatives, there is still a significant amount of energy in the spectrum for high and even ultrasonic frequencies. However, as we all know from using the (analog) phone, it seems that within a bandwidth of 4 kHz the speech signal contains all the information necessary to understand a human voice.

#### 5.1.2.2 Fundamental Frequency

As described earlier, using voiced excitation for the speech sound will result in a pulse train, the so-called fundamental frequency. Voiced excitation is used when articulating vowels and some of the consonants. For fricatives (e.g., /f/ as in fish or /s/, as in mess), unvoiced excitation (noise) is used. In these cases, usually no fundamental frequency can be detected. On the other hand, the zero crossing rate of the signal is very high. Plosives (like /p/ as in put), which use transient excitation, you can best detect in the speech signal by looking for the short silence necessary to build up the air pressure before the plosive bursts out.

### 5.1.3 A Very Simple Model of Speech Production

As we have seen, the production of speech can be separated into two parts: Producing the excitation signal and forming the spectral shape. Thus, we can draw a simplified model of speech production.

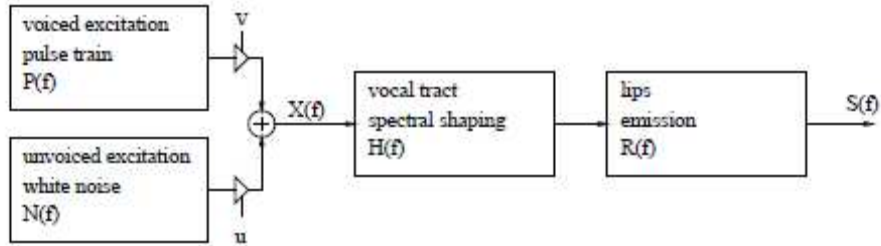

Fig.5.1: Model of a speech production

The spectrum S (f) of the speech signal is given as:

$$S(f) = (v \cdot P(f) + u \cdot N(f)) \cdot H(f) \cdot R(f) = X(f) \cdot H(f) \cdot R(f) \quad \ldots \quad (5.2)$$

To influence the speech sound, we have the following parameters in our speech production model:

• The mixture between voiced and unvoiced excitation (determined by v and u)

• The fundamental frequency (determined by P(f))

• The spectral shaping (determined by H(f))

• The signal amplitude (depending on v and u)

These are the technical parameters describing a speech signal. To perform speech recognition, the parameters given above have to be computed from the time signal (this is called speech signal analysis or "acoustic preprocessing") and then forwarded to the speech recognizer. For the speech recognizer, the most valuable information is contained in the way the spectral shape of the speech signal changes in time. To reflect these dynamic changes, the spectral shape is determined in short intervals of time, e.g., every 10 ms. By directly computing the spectrum of the speech signal, the fundamental frequency would be implicitly contained in the measured spectrum (resulting in unwanted "ripples" in the spectrum). Figure 5.2 shows the time signal of the vowel /a:/ and fig. 5.3 shows the logarithmic power spectrum of the vowel computed via FFT.

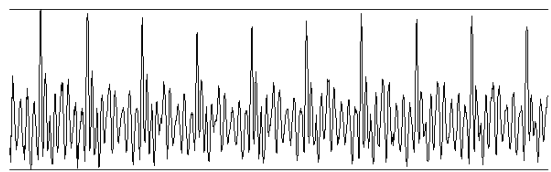

Fig. 5.2: Time signal of the vowel /a:/ (fs = 11kHz, length = 100ms).

The high peaks in the time signal are caused by the pulse train P (f) generated by voiced excitation.

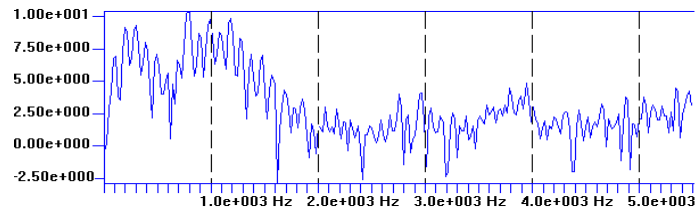

Fig. 5.3: Log power spectrum of the vowel /a:/ (fs = 11kHz, N = 512).

The ripples in the spectrum are caused by P(f). One could also measure the spectral shape by means of an analog filter bank using several bandpass filters as is depicted below. After rectifying and smoothing the filter outputs, the output voltage would represent the energy contained in the frequency band defined by the corresponding bandpass filter.

### 5.1.4 Speech Parameters Used by Speech Recognition Systems

As shown above, the direct computation of the power spectrum from the speech signal results in a spectrum containing "ripples" caused by the excitation spectrum X(f). Depending on the implementation of the acoustic preprocessing however, special transformations are used to separate the excitation spectrum X(f) from the spectral shaping of the vocal tract H(f). Thus, a smooth spectral shape (without the ripples), which represents H(f) can be estimated from the speech signal. Most speech recognition systems use the so–called MEL frequency CEPSTRAL coefficients (MFCC) and its first (and sometimes second) derivative in time to better reflect dynamic changes.

### 5.1.4.1 Computation of the Short Term Spectra

As we recall, it is necessary to compute the speech parameters in short time intervals to reflect the dynamic change of the speech signal. Typically, the spectral parameters of speech are estimated in time intervals of 10ms. First, we have to sample and digitize the speech signal. Depending on the implementation, a sampling frequency fs between 8kHz and 16kHz and usually a 16bit quantization of the signal amplitude is used. After digitizing the analog speech signal, we get a series of speech samples $s(k \cdot t)$ where $t = 1/fs$ or, for easier notation, simply $s(k)$.

Now a pre-emphasis filter is used to eliminate the -6dB per octave decay of the spectral energy.

$$\hat{s}(k) = s(k) - 0.97 \cdot s(k-1) \quad \ldots (5.2)$$

### 5.1.4.2 Mel Spectral Coefficients

As shown in perception experiments, the human ear does not show a linear frequency resolution but builds several groups of frequencies and integrates the spectral energies within a given group. Furthermore, the mid-frequency and bandwidth of these groups are non–linearly distributed. The non-linear warping of the frequency axis can be modeled by the so-called MEL scale. The frequency groups are assumed to be linearly distributed along the mel scale. The MEL–frequency fmel can be computed from the frequency f as follows:

$$\ldots (5.4)$$

$$f_{mel}(f) = 2595 \cdot \log\left(1 + \frac{f}{700Hz}\right)$$

Figure 5.4 shows a plot of the MEL scale. The human ear has high frequency resolution in low–frequency parts of the spectrum and low frequency resolution in the high–frequency parts of the spectrum. The coefficients of the power spectrum |V(n)|2 are now transformed to reflect the frequency resolution of the human ear. A common way to do this is to use K triangle–shaped windows in the spectral domain to build a weighted sum over those power spectrum coefficients |V(n)|2 which lie within the window. We denote the windowing coefficients as ηkn ; k = 0, 1, . . .K − 1 ; n = 0, 1, . . .N/2

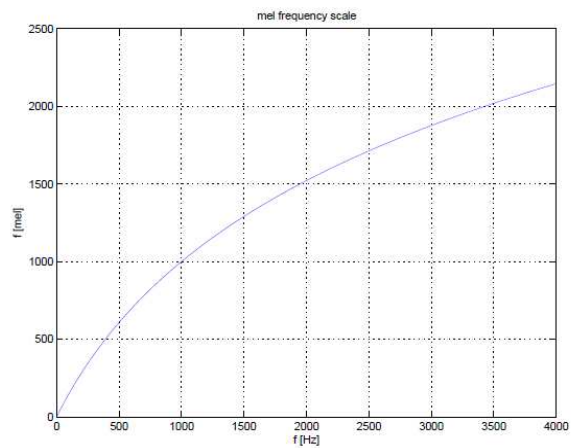

Fig. 5.4: The MEL Frequency scale

The so–called MEL spectral coefficients is given by,

$$G(k) = \sum_{n=0}^{N/2} \eta_{kn} \cdot |V(n)|^2 \; ; \; k = 0, 1, \ldots \mathcal{K} - 1 \qquad \ldots (5.5)$$

Caused by the symmetry of the original spectrum, the MEL power spectrum is also symmetric in k.

$$G(k) = G(-k) \qquad \ldots (5.6)$$

Therefore, it is sufficient to consider only the positive range of k.

### 5.1.4.3 Cepstral Transformation

Before we continue, let's remember how the spectrum of the speech signal was described by eq. 5.2. Since the transmission function of the vocal tract H(f) is multiplied with the spectrum of the excitation signal X(f), we had those unwanted "ripples" in the spectrum. For the speech recognition task, a smoothed spectrum is required which should represent H(f) but not X(f). To cope with this problem, cepstral analysis is used. If we look at eq. 5.2, we can separate the product of spectral functions into the interesting vocal tract spectrum and the part describing the excitation and emission properties:

$$S(f) = X(f) \cdot H(f) \cdot R(f) = H(f) \cdot U(f) \qquad \ldots (5.7)$$

We can now transform the product of the spectral functions to a sum by taking the logarithm on both sides of the equation:

$$\begin{aligned} \log(S(f)) &= \log(H(f) \cdot U(f)) \\ &= \log(H(f)) + \log(U(f)) \end{aligned} \qquad \ldots (5.8)$$

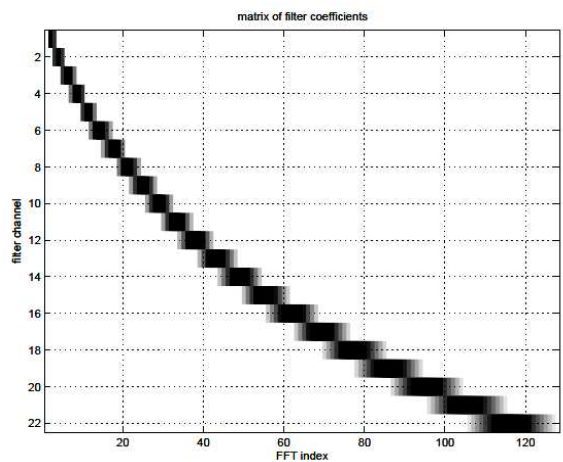

Fig. 5.5: Matrix of coefficients $\eta_{kn}$

This holds also for the absolute values of the power spectrum and also for their squares:

$$\begin{aligned}\log(|S(f)|^2) &= \log(|H(f)|^2 \cdot |U(f)|^2) \\ &= \log(|H(f)|^2) + \log(|U(f)|^2)\end{aligned} \quad \ldots (5.9)$$

In figure 5.2 we see an example of the log power spectrum, which contains unwanted ripples caused by the excitation signal $U(f) = X(f) \cdot R(f)$.

In the log–spectral domain we could now subtract the unwanted portion of the signal, if we knew $|U(f)|^2$ exactly. But all we know is that U(f) produces the "ripples", which now are an additive component in the log–spectral domain, and that if we would interpret this log–spectrum as a time signal, the "ripples" would have a "high frequency" compared to the spectral shape of $|H(f)|$. To get rid of the influence of U(f), one would have to get rid of the "high-frequency" parts of the log–spectrum (remember, we are dealing with the spectral coefficients as if they would represent a time signal). This would be a kind of low–pass filtering.

The filtering can be done by transforming the log–spectrum back into the time–domain (in the following, $FT^{-1}$ denotes the inverse Fourier transform):

$$\hat{s}(d) = \mathcal{FT}^{-1}\{\log(|S(f)|^2)\} = \mathcal{FT}^{-1}\{\log(|H(f)|^2)\} + \mathcal{FT}^{-1}\{\log(|U(f)|^2)\}$$

$$\ldots (5.10)$$

The ripples in the spectrum are caused by X(f). The inverse Fourier transforms brings us back to the time–domain (d is also called as delay), giving the so–called cepstrum (a reversed "spectrum"). The resulting cepstrum is real–valued, since $|U(f)|2$ and $|H(f)|2$ are both real-valued and both are even: $|U(f)|2 = |U(-f)|2$ and $|H(f)|2 = |H(-f)|2$. Applying the inverse DFT to the log power spectrum coefficients $\log(|V(n)|2)$ yields the following equation.

$$\hat{s}(d) = \frac{1}{N}\sum_{n=0}^{N-1}\log(|V(n)|^2) \cdot e^{j2\pi dn/N} \; ; \; d = 0, 1, \ldots N-1 \quad \ldots (5.11)$$

The Figure 5.6 shows the result of the inverse DFT applied on the log power spectrum shown in fig. 5.2.

The peak in the cepstrum reflects the ripples of the log power spectrum: Since the inverse Fourier transformation eq. 5.11 is applied to the log power spectrum, an oscillation in the frequency domain with a Hertz-period of

$$p_f = n \cdot \Delta f = n \cdot \frac{f_s}{N} = \frac{n}{N \cdot \Delta t} \qquad \ldots (5.12)$$

will show up in the cepstral domain as a peak at time t:

$$t = \frac{1}{p_f} = \frac{N}{n} \cdot \frac{1}{f_s} = \frac{N \cdot \Delta t}{n} \qquad \ldots (5.13)$$

The cepstral index d of the peak expressed as a function of the period length n is:

$$d = \frac{t}{\Delta t} = \frac{N \cdot \Delta t}{n \cdot \Delta t} = \frac{N}{n} \qquad \ldots (5.14)$$

The low–pass filtering of our energy spectrum can now be done by setting the higher-valued coefficients of s(d) to zero and then transforming back into the frequency domain. The process of filtering in the cepstral domain is also called *liftering*. In figure 5.7, all coefficients right of the vertical line were set to zero and the resulting signal was transformed back into the frequency domain, as shown in fig.5.8. One can clearly see that this results in a "smoothed" version of the log power spectrum if we compare figures 5.8 and 5.6.

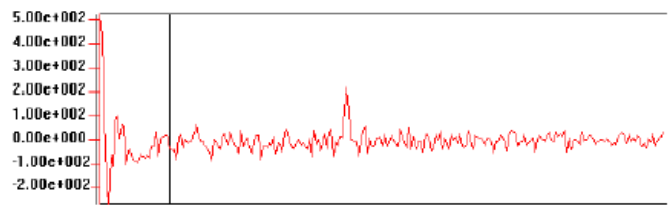

Fig. 5.6: Cepstrum of the vowel /a:/ (fs = 11kHz, N = 512).

The ripples in the spectrum result in a peak in the cepstrum.

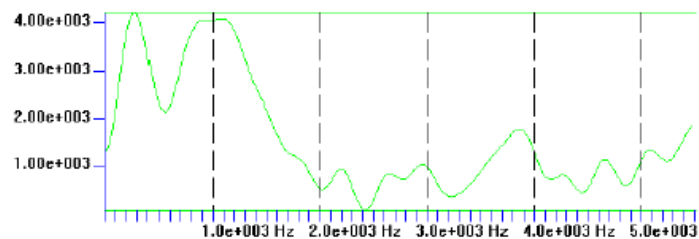

Fig. 5.7: Power spectrum of the vowel /a:/ after cepstral smoothing.

All but the first 32-cepstral coefficients were set to zero before transforming back into the frequency domain. It should be noted that due to the symmetry of |V(n)|2, is possible to replace the inverse DFT by the more efficient cosine transform:

$$\hat{s}(0) = \sqrt{\frac{2}{N}} \sum_{n=0}^{N/2-1} \log\left(|V(n)|^2\right) \qquad \ldots (5.15)$$

$$\hat{s}(d) = \sqrt{\frac{4}{N}} \sum_{n=0}^{N/2-1} \log\left(|V(n)|^2\right) \cdot \cos\left(\frac{\pi d(2n+1)}{N}\right) \; ; \; d = 1, 2, \ldots N/2 \qquad \ldots (5.16)$$

**5.1.4.4 MEL Cepstrum**

Now that we are familiar with the cepstral transformation and cepstral smoothing, we will compute the MEL cepstrum commonly used in speech recognition.

As stated above, for speech recognition, the MEL spectrum is used to reflect the perception characteristics of the human ear. In analogy to computing the cepstrum, we now take the logarithm of the MEL power spectrum (instead of the power spectrum itself) and transform it into the quefrency domain to compute the so–called MEL cepstrum. Only the first Q (less than 14) coefficients of the MEL cepstrum are used in typical speech recognition systems. The restriction to the first Q coefficients reflects the low–pass liftering process as described above.

Since the MEL power spectrum is symmetric due to eqn. (6), the Fourier-Transform can be replaced by a simple cosine transform:

$$c(q) = \sum_{k=0}^{K-1} \log\left(G(k)\right) \cdot \cos\left(\frac{\pi q(2k+1)}{2K}\right) \; ; \; q = 0, 1, \ldots, Q-1 \qquad \ldots (5.17)$$

While successive coefficients G(k) of the MEL power spectrum are correlated, the Mel Frequency Cepstral Coefficients (MFCC) resulting from the cosine transform eq. 5.17 are decorrelated. The MFCC are used directly for further processing in the speech recognition system instead of transforming them back to the frequency domain.

#### 5.1.4.5 Signal Energy

Furthermore, the signal energy is added to the set of parameters. It can simply be computed from the speech samples s(n) within the time window by:

$$e = \sum_{n=0}^{N-1} s^2(n)$$

... (5.18)

## 5.2 Dynamic Programming / Dynamic Time Warping

The procedure to compare two sequences of vectors is also known as Dynamic Programming (DP) or as Dynamic Time Warping (DTW). We will now define a more formal framework and will add some hints to possible realizations of the algorithm.

### 5.2.1 The Dynamic Programming Algorithm

In the following formal framework we will iterate through the matrix column by column, starting with the leftmost column and beginning each column at the bottom and continuing to the top. For ease of notation, we define d(i, j) to be the distance d($w_i$, $x_j$) between the two vectors $w_i$ and $x_i$.

Since we are iterating through the matrix from left to right, and the optimization for column j uses only the accumulated distances from columns j and j−1, we will use only two arrays $δ_j(i)$ and $δ_{j−1}(i)$ to hold the values for those two columns instead of using a whole matrix for the accumulated distances δ(i, j): Let $δ_j(i)$ be the accumulated distance δ(i, j) at grid point (i, j) and $δ_{j−1}(i)$ the accumulated distance δ(i, j − 1) at grid point (i, j − 1).

It should be mentioned that it possible to use a single array for time indices j and j − 1. One can overwrite the old values of the array with the new ones.

To keep track of all the selections among the path hypotheses during the optimization, we have to store each path alternative chosen for every grid point. We could for every grid point (i, j) either store the indices k and l of the predecessor point (k, l) or we could only store a code number for one of the three path alternatives (horizontal, diagonal and vertical path) and compute the predecessor point (k, l) out of the code and the current point (i, j).

For ease of notation, we assume in the following that the indices of the predecessor grid point will be stored: Let ψ(i, j) be the predecessor grid point (k, l) chosen during the optimization step at grid point (i, j).

With these definitions, the dynamic programming (DP) algorithm can be written as follows:

**Initialization:**

◦ Grid point (0, 0):

$$\psi(0, 0) = (-1, -1)$$

$$\delta_j(0) = d(0, 0) \quad \quad \ldots \quad (5.19)$$

◦ Initialize first column (only vertical path possible):

for i = 1 to TW − 1

{ $\delta_j(i) = d(i, 0) + \delta_j(i - 1)$

$\psi(i, 0) = (i - 1, 0)$ … (5.20)

}

**Iteration:**

for j = 1 to TX − 1  // Compute all columns:

{ ◦ swap arrays $\delta_{j-1}(.)$ and $\delta_j(.)$

◦ first point (i = 0, only horizontal path possible):

$\delta_j(0) = d(0, j) + \delta_{j-1}(0)$

$\psi(0, j) = (0, j - 1)$ … (5.21)

◦ compute column j:

for i = 1 to TW − 1

{ ◦ optimization step:

$$\delta_j(i) = \min \begin{cases} \delta_{j-1}(i) & + & d(i, j) \\ \delta_{j-1}(i-1) & + & 2 \cdot d(i, j) \\ \delta_j(i-1) & + & d(i, j) \end{cases} \quad \ldots \quad (5.22)$$

◦ tracking of path decisions:

$$\psi(i,j) = \underset{(k,l) \in \left\{\begin{array}{c}(i,j-1),\\(i-1,j-1),\\(i-1,j)\end{array}\right\}}{\arg\min} \left\{\begin{array}{ccc}\delta_{j-1}(i) & + & d(i,j)\\\delta_{j-1}(i-1) & + & 2 \cdot d(i,j)\\\delta_{j}(i-1) & + & d(i,j)\end{array}\right\} \quad \ldots (5.23)$$

}
}

**Termination:**

$$D(TW - 1, TX - 1) = \delta_j(TW - 1, TX - 1) \quad \ldots (5.24)$$

**Backtracking:**

◦ Initialization: i = TW − 1, j = TX − 1  … (5.25)

◦ while ψ(i, j) 6= −1

◦ get predecessor point:  i, j = ψ(i, j)  … (5.26)

**Additional Constraints:**

Note that the algorithm above will find the optimum path by considering all path hypotheses through the matrix of grid points. However, there will be paths regarded during computation, which are not very likely to be the optimum path due to the extreme time warping they involve. For example, think of the borders of the grid: There is a path going horizontally from (0, 0) until (TW − 1, 0) and then running vertically to (TW − 1, TX − 1).

Paths running more or less diagonally trough the matrix will be much more likely candidates for being the optimal path. To avoid unnecessary computational efforts, the DP algorithm is therefore usually restricted to searching only in a certain region around the diagonal path leading from (0, 0) to (TW −1, TX −1).

Of course, this approach takes the (small) risk of missing the globally optimal path and instead finding only the best path within the defined region. Later we will learn far more sophisticated methods to limit the so–called search space of the DP algorithm.

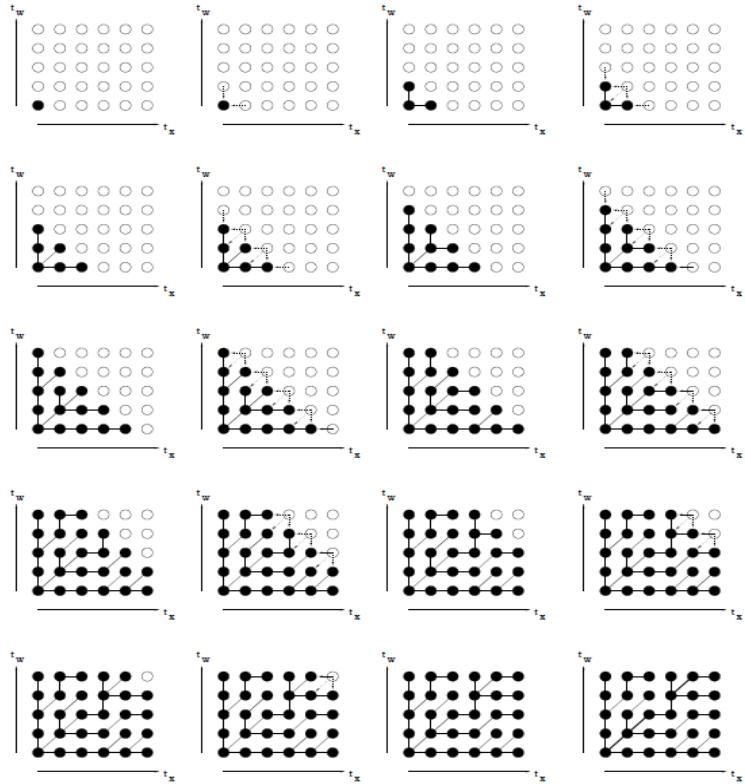

Fig. 5.8: Iteration steps finding the optimal path

## 5.3 Recognition of Isolated Words

Computing the distance between an unknown vector sequences to all prototypes of all classes and then assigning the unknown sequence to the class having the smallest class distance perform the classification of words. Now we will reformulate the classification task so that it will depend directly on the optimum path chosen by the dynamic programming algorithm and not by the class distances.

While the description of the DTW classification algorithm in chapter 4 might let us think that one would compute all the distances sequentially and then select the minimum distance, it is more useful in practical applications to compute all the distances between the unknown vector sequence and the class prototypes in parallel. This is possible since the DTW algorithm needs only the values for time index t and (t−1) and therefore there is no need to wait until the utterance of the unknown vector sequence is completed. Instead, one can start with the recognition process immediately as soon as the utterance begins (we will not deal with the question of how to recognize the start and end of an utterance here). To do so, we have to reorganize our search space a little bit.

First, lets assume the total number of all prototypes over all classes is given by M. If we want to compute the distances to all M prototypes simultaneously, we have to keep track of the accumulated distances between the unknown vector sequence and the prototype sequences individually.

Hence, instead of the column (or two columns, depending on the implementation) we used to hold the accumulated distance values for all grid points, we now have to provide M columns during the DTW procedure. Now we introduce an additional "virtual" grid point together with a specialized local path alternative for this point: The possible predecessors for this point are defined to be the upper–right grid points of the individual grid matrices of the prototypes. In other words, the virtual grid point can only be reached from the end of each prototype word, and among all the possible prototype words, the one with the smallest accumulated distance is chosen. By introducing this virtual grid point, the classification task itself (selecting the class with the smallest class distance) is integrated into the framework of finding the optimal path.

Now all we have to do is to run the DTW algorithm for each time index j and along all columns of all prototype sequences. At the last time slot (TW − 1) we perform the optimization step for the virtual grid point, i.e., the predecessor grid point to the virtual grid point is chosen to be the prototype word having the smallest accumulated distance.

Note that the search space we have to consider is spanned by the length of the unknown vector sequence on one hand and the sum of the length of all prototype sequences of all classes on the other hand. Figure 5.10 shows the individual grids for the prototypes (only three are shown here) and the selected optimal path to the virtual grid point.

The backtracking procedure can of course be restricted to keeping track of the final optimization step when the best predecessor for the virtual grid point is chosen. Assigning the unknown vector sequence to the very class to which the prototype belongs whose word end grid point was chosen then performs the classification task.

Only a verbal description was given and we did not bother with a formal description.

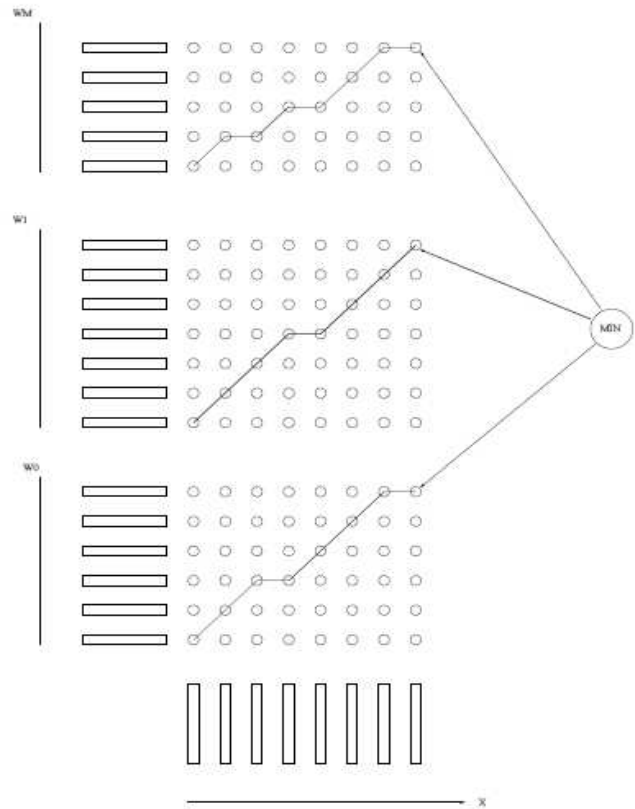

Fig.5.9: Classification task redefined as finding the optimal path among all prototype words

However, by the reformulation of the DTW classification we learned a few things:

• The DTW algorithm can be used for real–time computation of the distances

• The classification task has been integrated into the search for optimal path

• Instead of the accumulated distance, now the optimal path itself is important for the classification task.

Hence, Using the DTW algorithm, we have written a program to find a match between spoken word and previously stored template of words. The results of this algorithm are shown in the next chapter.

# CHAPTER 6

# RESULTS

In this chapter we have presented various promising results for this project. The first section shows the results of speech processing module. The next two sections show the results of obstacle detection module with calibration and performance of the system in real time. The voice recognition module gave 80% accuracy in identifying words. The obstacle detection module gave more than 90% accuracy in estimating the distance of the obstacles.

## 6.1 Results of Voice recognition

The Matlab program for speech recognition (figure 6.1) is run for various speech command inputs and the results for all words are given below.

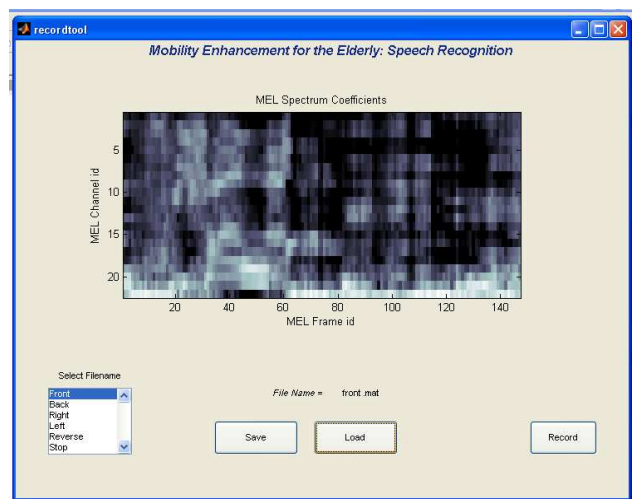

Fig. 6.1: MEL Spectrum for the speech input of word "Front"

The matlab command window output display for the speech inputs "Front", "Back", "Right" and "Left" and "stop" are shown below.

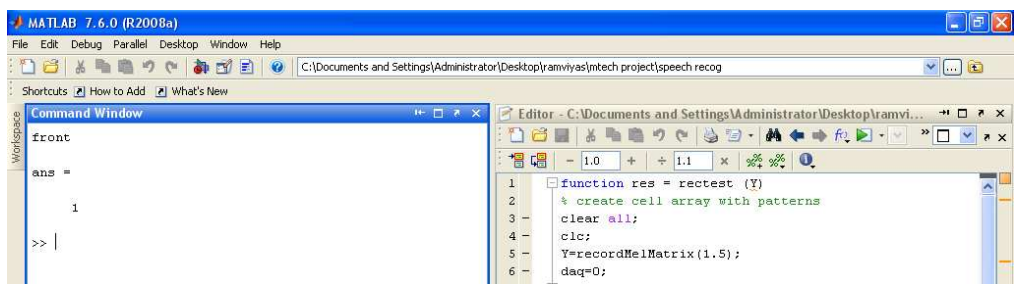

Fig. 6.2: Matlab Command Window Output for the speech input – Front

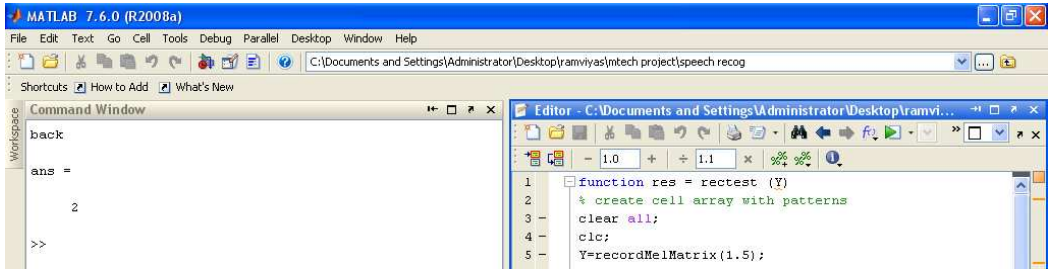

Fig. 6.3: Matlab Command Window Output for the speech input – Back

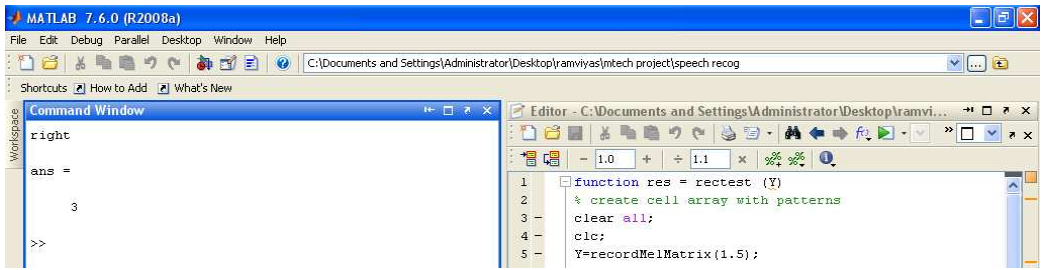

Fig. 6.4: Matlab Command Window Output for the speech input – Right

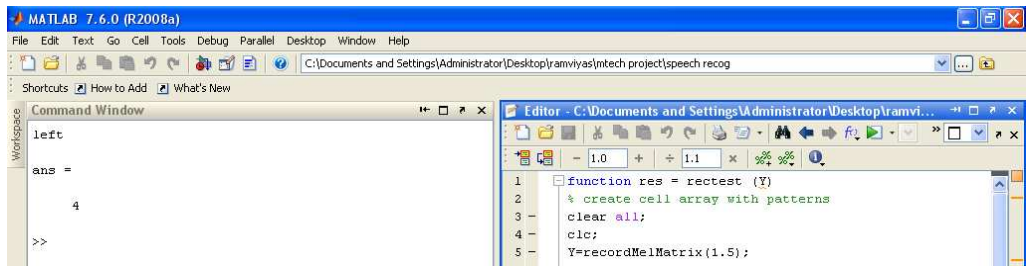

Fig. 6.5 Matlab Command Window Output for the speech input – Left

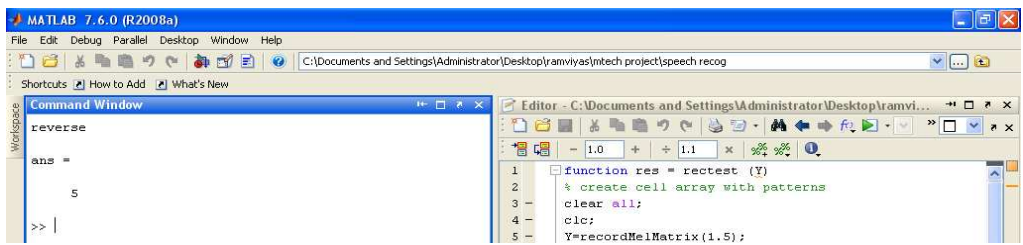

Fig. 6.5: Matlab Command Window Output for the speech input – Reverse

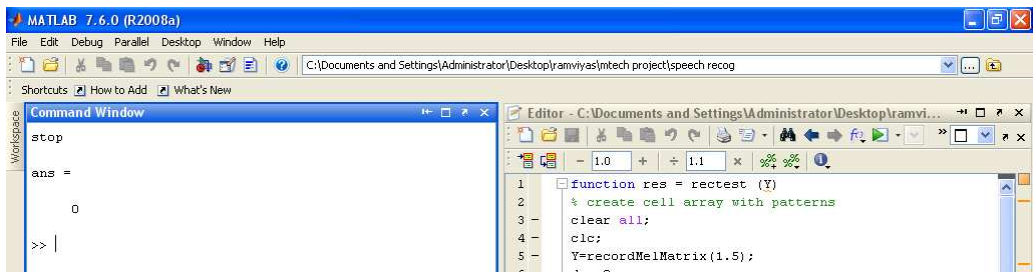

Fig. 6.6: Matlab Command Window Output for the speech input – Stop

Having conducted a survey with various volunteers, accuracy of response lies between **75-80%** which if the algorithm is improved, can yield up to 90% accuracy and will be sufficient for a wheelchair application. This module can be very useful to a person who is not able to use his/her hands freely due to handicap or the age problem. The user can give voice commands to control the movement of the wheelchair without using his hands for controlling joystick or mouse. This module can be enhanced not only to control the wheelchair but also to provide various other functionalities such as getting reminders like time, place, date, etc.

## 6.2 Calibration results for stereo vision setup

The main purpose of calibration is to find the focal length f of the camera and the baseline distance T between the two cameras. The error obtained in the calibration procedure should be less than +/-0.5 pixels.

We have taken pictures of a standard chessboard pattern in different orientations and run the Zhang's calibration algorithm using [27]. The input images are shown in Figure 6.7.

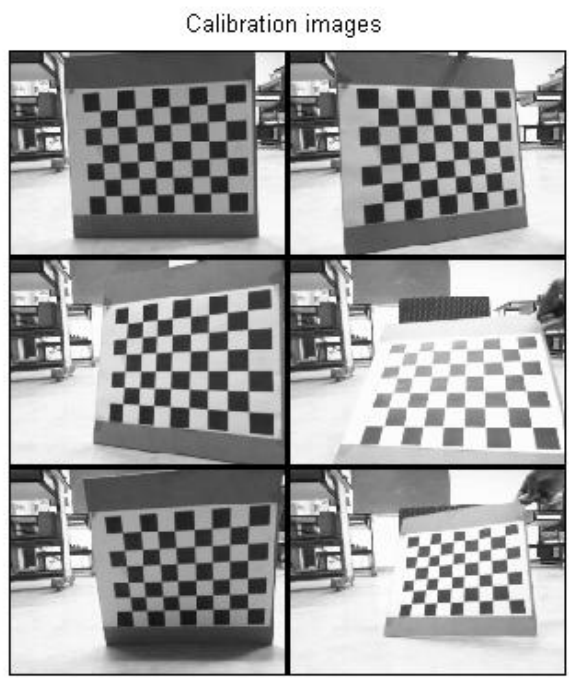

Figure 6.7: Calibration input images

Then, we have run the calibration process for both right and left images and then the stereo calibration is executed. The result of stereo calibration is shown below. We can observe from the results that the focal length of both the cameras is around 300mm for 320x240 pixel resolution in the Microsoft Lifecam Cinema camera. In the results shown in figure 6.8, we can observe that the focal length f is around 300 (Pixel units). The distortion values are almost zero which means the errors are negligible.

```
Stereo calibration parameters:

Intrinsic parameters of left camera:

Focal Length:          fc_left = [ 296.01142   296.62390 ] ± [ 3.36876   3.22968 ]
Principal point:       cc_left = [ 155.75326   114.41865 ] ± [ 3.70888   2.54002 ]
Skew:             alpha_c_left = [ 0.00000 ] ± [ 0.00000 ]  => angle of pixel axes = 90.00000 ± 0.00000 degrees
Distortion:            kc_left = [ 0.02002   -0.07350   -0.00172   -0.01327  0.00000 ] ± [ 0.02822   0.06960   0.00257   0.00466  0.00000 ]

Intrinsic parameters of right camera:

Focal Length:         fc_right = [ 298.26905   299.67879 ] ± [ 3.38423   3.28076 ]
Principal point:      cc_right = [ 147.58032   112.59891 ] ± [ 3.78957   2.72342 ]
Skew:            alpha_c_right = [ 0.00000 ] ± [ 0.00000 ]  => angle of pixel axes = 90.00000 ± 0.00000 degrees
Distortion:           kc_right = [ 0.01915   -0.07523   -0.00025   -0.01785  0.00000 ] ± [ 0.04284   0.18778   0.00329   0.00440  0.00000 ]

Extrinsic parameters (position of right camera wrt left camera):

Rotation vector:     om = [ -0.00686   -0.02761   -0.00880 ] ± [ 0.01040   0.01608   0.00091 ]
Translation vector:   T = [ -40.85256   1.78956   3.50527 ] ± [ 0.80405   0.72203   3.55755 ]

Note: The numerical errors are approximately three times the standard deviations (for reference).
```

Figure 6.8: Calibration results

The values in the rotational vector are also near to zero signifies that the orientation of both the cameras are similar. The translational vector shows -40.85 mm as X axis, 1.7mm for Y axis and 3.5mm for Z axis. These values show the linear displacement between the cameras in 3 dimensions. The displacement in Y and Z axis are negligible. Hence, the baseline distance T is found to be 40.85mm which we use along with f and disparity value for calculating the distance value using eq. 4.1.

## 6.3 Run time Performance of obstacle detection system

The obstacles are kept at various distances and the output is recorded which are shown in the following figures.

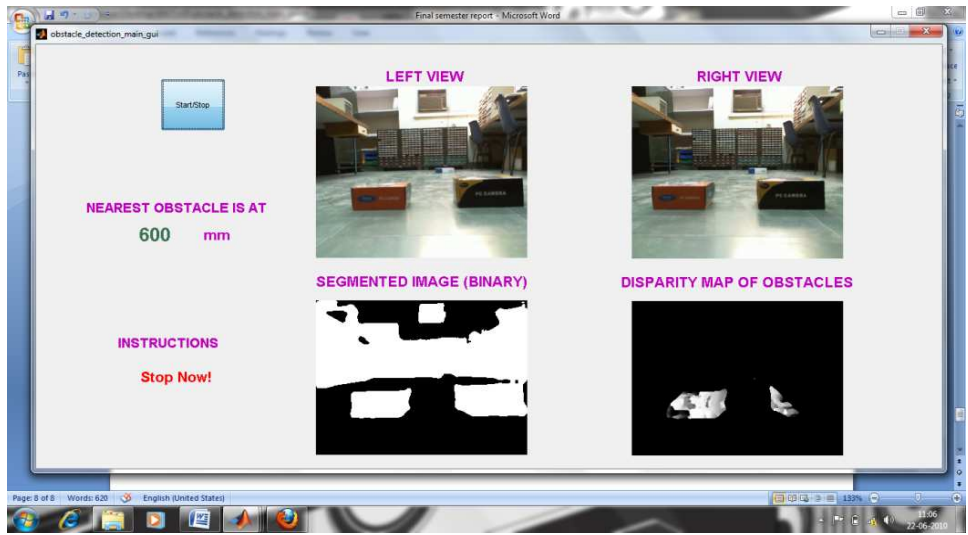

Figure 6.9: Obstacle at 600mm

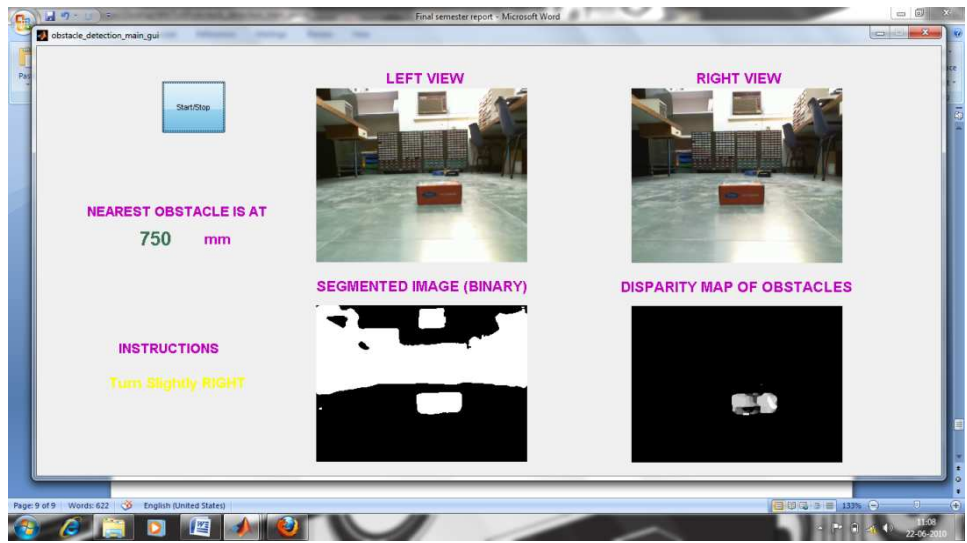

Figure 6.10: Obstacle at 750mm

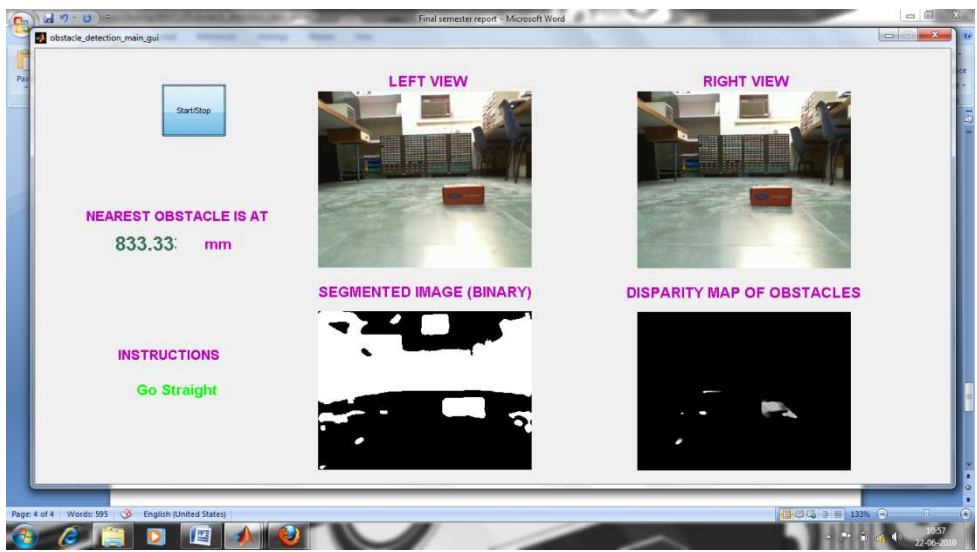

Figure 6.11: Obstacle at 830mm

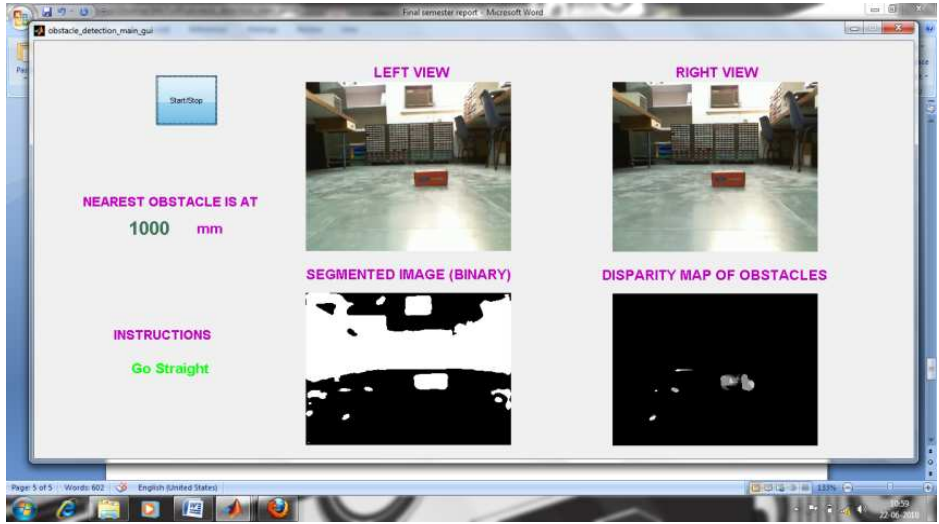

Figure 6.12: Obstacle at 1000mm

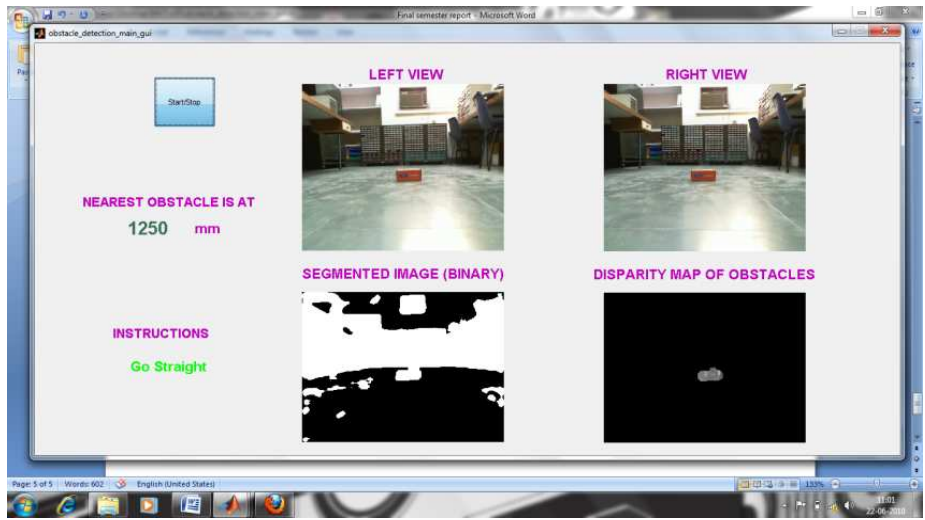

Figure 6.13: Obstacle at 1300mm

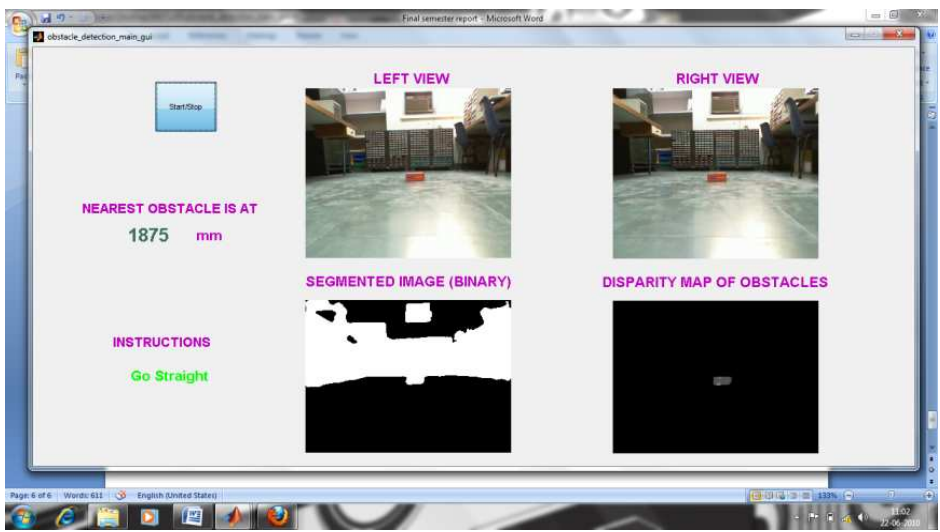

Figure 6.14: Obstacle at 1850mm

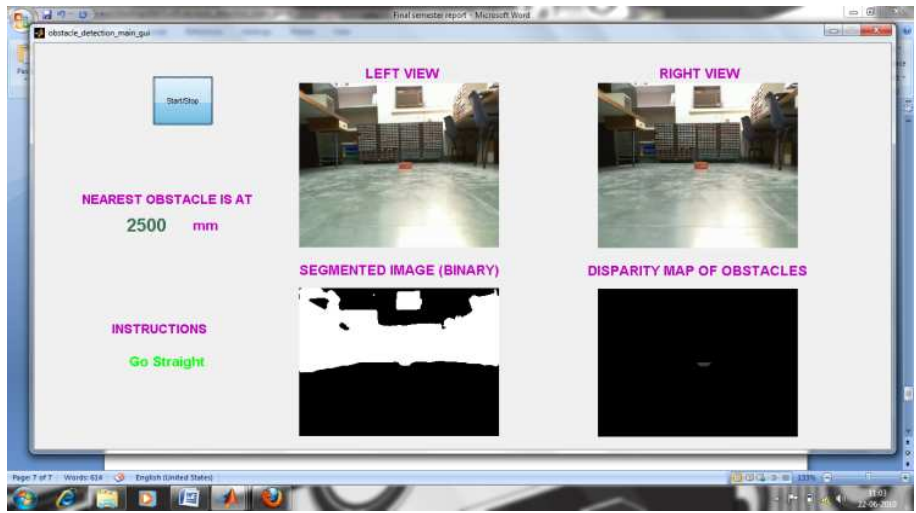
Figure 6.15: Obstacle at 2500mm

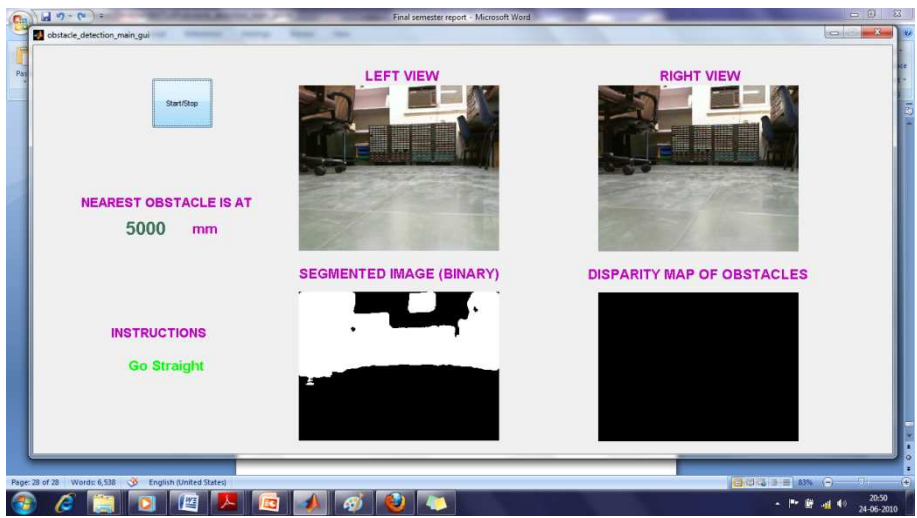
Figure 6.16: No Obstacles in the path

The program is written in such a way that, if there are no obstacles, then the distance of nearest obstacle will be 5000mm. In the figure 6.10, since the obstacle distance is between 600 and 750mm, it shows the instruction as Turn slightly Right based on the space available on both sides. Also, in figure 6.9, since the obstacle is very close, the Stop instruction is given.

Various experiments have been performed and the readings are noted for range and linearity of the output. The graph in figure 6.17 shows the actual and measured distances of the obstacle.

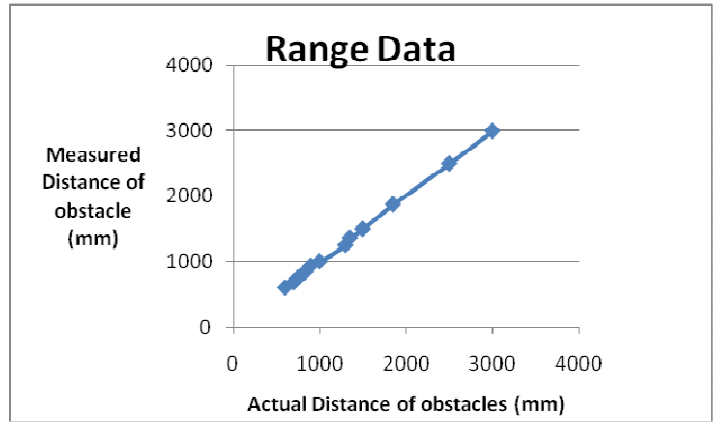

Figure 6.17: Range data for accuracy analysis

From the repeated trails, we have observed that the average error over a range of 600mm to 3000mm is **15mm**. The time taken for the each iteration is 1 second in Intel Core2Duo 2.1GHz processor with 4GB RAM. The Matlab version used is R2009a running in Windows 7 Operating system.

By assuming the wheelchair is moving at a rate of 2000m/hr (equivalent to half the walking speed), then in 1 second, the chair can move approx 1.2m. Therefore, we have set the threshold value to stop the wheelchair if the obstacle is < 0.6m ahead. Also, we must analyze the range data up to 2m to 4m range so that obstacle data up to 2m range is available always. The threshold for stopping the wheelchair/robot can be changed according to the speed.

Thus this module can be incorporated in a wheelchair for automatically controlling the movement of the wheelchair to avoid obstacles in the path and also to find the path to navigate without avoiding obstacles, if possible. This facility will be useful to the aged people or people who are having low vision capability which makes them poor in visual judgement to see obstacles. Also, this module can be enhanced to use image processing features for identifying thee visitors through face recognition, identifying the location using some features in the image like locating the doors of the room, etc.

Since we have implemented in Matlab, the time taken is more. It is assumed that, if we implement the same algorithm in C program, then the time taken will be reduced to at least half. Presently we have used video resolution of 320x240 pixels in both the cameras. If we decrease the resolution of the video input, then the time taken will be reduced accordingly.

# CHAPTER 7

# CONCLUSION

## 7.1 Summary

This thesis introduces a new robust method for obstacle detection in an unknown environment for Robotic applications, particularly for mobility enhancement for the elderly people using wheelchair with enhanced robotic facilities. We have also presented the Speech recognition part with successful implementation and results. First, we presented the motivation and literature survey to define the problem in more details. Then, we discussed the principles and approach to achieve the solution in chapter 3 to 5.

We encountered problems with lighting conditions and mechanical alignment of the setup. By optimizing the Exposure and brightness settings of the camera, we have reduced the problems due to illumination changes. Several experiments have been conducted to set various threshold values.

Optimizing the Baseline distance between cameras also important as wide baseline involves more computation but good resolution and small baseline gives poor resolution but easy computation. The results shown in chapter 5 are promising and the system works well in real time environment. The obtained results are

- ✓ Range: 2500mm
- ✓ Resolution: 25mm
- ✓ Accuracy: 15mm over 600mm to 3000mm range
- ✓ Speed of analysis: 1 sec.

## 7.2 Future work

The current method uses resolution of 320x240pixels. By reducing the resolution and implementing the algorithm in C/C++ language can improve the time performance of the system. Also, Dynamic programming approach for finding disparity map can perform better in both time and accuracy improvement. Furthermore, we can further improve the system to find the best path for navigation.

# Appendix

## A. Source code – Obstacle detection

### A.1 Main program

```matlab
function varargout = obstacle_detection_main_gui(varargin)
gui_Singleton = 1;
gui_State = struct('gui_Name',       mfilename, ...
                   'gui_Singleton',  gui_Singleton, ...
                   'gui_OpeningFcn', @obstacle_detection_main_gui_OpeningFcn, ...
                   'gui_OutputFcn',  @obstacle_detection_main_gui_OutputFcn, ...
                   'gui_LayoutFcn',  [] , ...
                   'gui_Callback',   []);
if nargin && ischar(varargin{1})
    gui_State.gui_Callback = str2func(varargin{1});
end

if nargout
    [varargout{1:nargout}] = gui_mainfcn(gui_State, varargin{:});
else
    gui_mainfcn(gui_State, varargin{:});
end

function obstacle_detection_main_gui_OpeningFcn(hObject, eventdata, handles, varargin)
handles.output = hObject;

guidata(hObject, handles);

function varargout = obstacle_detection_main_gui_OutputFcn(hObject, eventdata, handles)

varargout{1} = handles.output;

function togglebutton1_Callback(hObject, eventdata, handles)
global vid;
```

```matlab
if(get(hObject,'Value')==1)
    vid=plotData(handles);
    for i=1:50
        if(get(hObject,'Value')==0)
            break;
        end
        [HF dispMapSc depth]=obstacle_new_ncc(vid);
        axes(handles.axes4);
        imshow(HF);
        axes(handles.axes6);
        imshow(dispMapSc);
        set(handles.text6, 'String', num2str(depth));
        set(handles.text11, 'String','Go Straight','ForegroundColor','GREEN');
        B=zeros(1,320);
        if(depth<=600)
            %disp('Stop Now!');
             set(handles.text11, 'String','Stop Now!','ForegroundColor','RED');
        else if (depth<=750)
                for j=1:320
                    A=find(HF(1:210,j)==1);
                    if(~isempty(A))
                        B(j)=A(end);
                    end
                end
                R=[mean(B(1:107)) mean(B(108:214)) mean(B(215:320))];
                [val RR]=min(R(1:2:3));
                if RR==1
                    %disp('Turn slightly LEFT');
                    set(handles.text11, 'String','Turn Slightly LEFT','ForegroundColor','Yellow');
                else
                    %disp('Turn slightly RIGHT');
                    set(handles.text11, 'String','Turn Slightly RIGHT','ForegroundColor','Yellow');
                end
            end
        end

    end
```
58

```matlab
else
    stoppreview(vid);
end

function vid = plotData(handles)
axes(handles.axes3);
vid(1)=videoinput('winvideo',1,'YUY2_320x240');
vid(2)=videoinput('winvideo',2,'YUY2_320x240');
vid.ReturnedColorSpace='RGB';
vidres=vid(1).VideoResolution;
nBands=vid(1).NumberOfBands;
src1=getselectedsource(vid(1));
src2=getselectedsource(vid(2));
src1.FocusMode='manual';
src2.FocusMode='manual';
src1.Exposure=0;
src2.Exposure=0;
src1.Brightness=100;
src2.Brightness=100;
hImage = image(zeros(vidres(2),vidres(1), nBands),'Parent',handles.axes3);
preview(vid(1), hImage);
hImage = image(zeros(vidres(2),vidres(1), nBands),'Parent',handles.axes5);
preview(vid(2), hImage);
```

## A.2 Subroutine

```matlab
function [HF dispMapSc depth] = obstacle_new_ncc(vid)
 f=im2double(ycbcr2rgb([getsnapshot(vid(1)) getsnapshot(vid(2))]));
 IL=f(:,1:320,:);
 IR=f(:,321:640,:);
HL=rgb2hsv(IL);
H2 = imcrop(HL,[20.0000  180.0000  280.0000  60.0000]);
HV=HL(:,:,3);
HH=HL(:,:,1);
H2V=H2(:,:,3);
H2S=H2(:,:,2);
H2H=H2(:,:,1);
[k l]=size(H2V);
%bins=5;
HF=zeros(k,l);
```

```matlab
for i=1:k
    for j=1:l
        if (H2V(i,j)<=0.05 || H2S(i,j)<=0.1)
            H2H(i,j)=2;
        end
    end
end

HistrefH= sum((histc(H2H,[0 0.2001 0.4001 0.6001 0.8001 1])),2);
HistrefV= sum((histc(H2V,[0 0.2001 0.4001 0.6001 0.8001 1])),2);
HistrefH(5)=HistrefH(5)+HistrefH(6);
HistrefV(5)=HistrefV(5)+HistrefV(6);

for i=1:size(HV,1)
    for j=1:size(HV,2)
        bin_val=bin(HV(i,j));
        bin_hue=bin(HH(i,j));
         if
(HistrefV(bin_val)<=max(HistrefH)/50)||(HistrefH(bin_hue)<=max(HistrefV)/50
)           HF(i,j)=1;
        else HF(i,j)=0;
        end
    end
end

HF=medfilt2(HF, [9 9]);
bl=50; %baseline between cameras is 50mm
f=300; %focal length is 300pixels for 320x240 pixel size image
depth=5000;
nD=25; %Max disparity 30 pixels to give atleast 500m gap!
k1=0;
dispMap=zeros(240,320);
dispMin=0;
dispMax=nD;
win=(9-1)/2;
leftImage=rgb2gray(IL);
rightImage=rgb2gray(IR);
for i=120:210 %i=(1+win):(240-win) %%Row
    %q=2;
    for j=100-k1:200+k1 %j=(1+win):(320-nD-win) % %Column
```

```matlab
            if(HF(i,j)==1)
            prevNCC = 0.0;
            bestMatchSoFar = dispMin;
            for dispRange=dispMin:1:dispMax
                %ncc=0.0;
                nccNumerator=0.0;
                %nccDenominator=0.0;
                nccDenominatorRightWindow=0.0;
                nccDenominatorLeftWindow=0.0;
                for a=-win:1:win
                    for b=-win:1:win

nccNumerator=nccNumerator+(rightImage(i+a,j+b)*leftImage(i+a,j+b+dispRange));

nccDenominatorRightWindow=nccDenominatorRightWindow+(rightImage(i+a,j+b)*rightImage(i+a,j+b));

nccDenominatorLeftWindow=nccDenominatorLeftWindow+(leftImage(i+a,j+b+dispRange)*leftImage(i+a,j+b+dispRange));
                    end
                end

nccDenominator=sqrt(nccDenominatorRightWindow*nccDenominatorLeftWindow);
                ncc=nccNumerator/nccDenominator;
                if (prevNCC < ncc)
                    prevNCC = ncc;
                    bestMatchSoFar = dispRange;
                end
            end
            dispMap(i,j) = bestMatchSoFar;
            end
        end
        k1=k1+1;
end

dispMap=medfilt2(dispMap, [5 5]);
%figure; imshow(dispMap/nD);
```

```matlab
%x=max(max(op))*nD;
%y=hist(max(op)*nD,x);
x=1:nD;
y=sum(histc(dispMap,x),2);
for i=nD:-1:1
    if(y(i)>(3*nD)) % threshold is nD
        depth=bl*f/i;
        break;
    end
end
dispMapSc=dispMap/nD;
```

## B. Source code – Speech Recognition
### B.1 Main Program (rectest.m)

```matlab
function res = rectest (Y)
Y=recordMelMatrix(1.5);      % create cell array with patterns
daq=0;
for i = 1:6
    filename=filenamedet(i,daq);
    load (filename,'X');
    P{i} = X;
end
filename = 'stop';     % last pattern is zero pattern
load (filename,'X');
P{6} = X;
%compare patterns with Y
res = 0;
DP = dp_asym(P{6},Y);
min = DP.dist;
for i = 1:6
    DP = dp_asym(P{i},Y);
    t = DP.dist;
    if (t < min)
        res = i;
        min = t;
    end
end
%Output to DAQ card
daq=1;
dio=digitalio('nidaq','Dev1');
addline(dio,0:3,1,'out');
putvalue(dio,0);
disp(filenamedet(res,daq,dio)); %display the result
function out = filenamedet(number,daq,dio)
switch number
    case 1
        out='front';
        if(daq==1) putvalue(dio,1); %Glow LED for Front
        end
    case 2
        out='back';
        if(daq==1) putvalue(dio,2); %Glow LED for Back
        end
```

```matlab
    case 3
        out='right';
        if(daq==1) putvalue(dio,4); %Glow LED for Right
        end
    case 4
        out='left';
        if(daq==1) putvalue(dio,8); %Glow LED for Left
        end
    case 5
        out='reverse';
    case 6
        out='stop';
    otherwise
        out='stop';
 end;
```

## B.2.Training Program (recordtool.m)

```matlab
function varargout = recordtool(varargin)
gui_Singleton = 1;
gui_State = struct('gui_Name',       mfilename, ...
                   'gui_Singleton',  gui_Singleton, ...
                   'gui_OpeningFcn', @recordtool_OpeningFcn, ...
                   'gui_OutputFcn',  @recordtool_OutputFcn, ...
                   'gui_LayoutFcn',  [] , ...
                   'gui_Callback',   []);
if nargin & isstr(varargin{1})
    gui_State.gui_Callback = str2func(varargin{1});
end

if nargout
    [varargout{1:nargout}] = gui_mainfcn(gui_State, varargin{:});
else
    gui_mainfcn(gui_State, varargin{:});
end
% --- Executes just before recordtool is made visible.
function recordtool_OpeningFcn(hObject, eventdata, handles, varargin)
% This function has no output args, see OutputFcn.
% hObject    handle to figure
% eventdata  reserved - to be defined in a future version of MATLAB
% handles    structure with handles and user data (see GUIDATA)
% varargin   command line arguments to recordtool (see VARARGIN)
% Choose default command line output for recordtool
handles.output = hObject;
%set listbox
set(handles.File,'Min',1,'Max',6);
%init params
handles.time = 1.5;
handles.X = zeros(1,1);
handles.number = 1;
%handles.filename = sprintf('pattern_%d',handles.number);
handles.filename=filenamedet(handles.number);
set(handles.stext,'String',handles.filename)
colormap(bone);
% Update handles structure
guidata(hObject, handles);
% UIWAIT makes recordtool wait for user response (see UIRESUME)
function varargout = recordtool_OutputFcn(hObject, eventdata, handles)
% varargout  cell array for returning output args (see VARARGOUT);
% hObject    handle to figure
% eventdata  reserved - to be defined in a future version of MATLAB
```

```matlab
% handles    structure with handles and user data (see GUIDATA)
% Get default command line output from handles structure
varargout{1} = handles.output;
% --- Executes on button press in recordbutton.
function recordbutton_Callback(hObject, eventdata, handles)
% hObject    handle to recordbutton (see GCBO)
% eventdata  reserved - to be defined in a future version of MATLAB
% handles    structure with handles and user data (see GUIDATA)
handles.X = recordMelMatrix(handles.time);
guidata(hObject, handles);
plotData(handles);
% --- Executes on button press in savebutton.
function savebutton_Callback(hObject, eventdata, handles)
% hObject    handle to savebutton (see GCBO)
% eventdata  reserved - to be defined in a future version of MATLAB
% handles    structure with handles and user data (see GUIDATA)
X = handles.X;
save (handles.filename, 'X');
% --- Executes during object creation, after setting all properties.
function File_CreateFcn(hObject, eventdata, handles)
if ispc
    set(hObject,'BackgroundColor','white');
else

set(hObject,'BackgroundColor',get(0,'defaultUicontrolBackgroundColor'));
end
% --- Executes on selection change in File.
function File_Callback(hObject, eventdata, handles)
number = get(handles.File,'value');
handles.filename=filenamedet(number);
set(handles.stext,'String',handles.filename)
guidata(hObject, handles);
function plotData(handles)
axes(handles.axes1);
imagesc(flipud(handles.X));
xlabel('MEL Frame id');
ylabel('MEL Channel id');
title('MEL Spectrum Coefficients');
% --- Executes on button press in load.
function load_Callback(hObject, eventdata, handles)
load (handles.filename, 'X');
handles.X = X;
guidata(hObject, handles);
plotData(handles);
function out = filenamedet(number)
switch number
    case 6
        out='stop';
    case 1
        out='front';
    case 2
        out='back';
    case 3
        out='right';
    case 4
        out='left';
    case 5
        out='reverse';
    otherwise
        out='none';
end;
```

### B.3 Sub-function recordMelMatrix.m

```matlab
function X = recordMelMatrix(sec)
data.T = sec; %seconds to record
data.fs = 8000; %sampling frequency
data.samples = data.T * data.fs; %number of samples to record
data.N=256; %fft length
data.shift = 10*data.fs/1000; %time shift is 10ms
data.nofChannels = 22; %number of mel filter channels
%compute matrix of mel filter coefficients
data.W = melFilterMatrix(data.fs,data.N,data.nofChannels);
data.s = wavrecord(data.samples,data.fs,'double');
wavplay(data.s,data.fs);

%compute mel spectra
data.MEL = computeMelSpectrum(data.W,data.shift,data.s);
data.nofFrames = size(data.MEL.M,2);
data.nofMelChannels = size(data.MEL.M,1);
%normalize energy of mel spectra
%take log value
epsilon = 10e-5;
for k = 1:data.nofFrames
    for c = 1:data.nofMelChannels
        %normalize energy
        data.MEL.M(c,k) = data.MEL.M(c,k)/data.MEL.e(k);
        %take log energy
        data.MEL.M(c,k) = loglimit(data.MEL.M(c,k),epsilon);
    end %for c
end %for k
X = data.MEL.M ;
```

### B.4 Sub-function computeMelSpectrum.m

```matlab
function MEL = computeMelSpectrum (W,winShift,s)
% compute local variables
[nofChannels,maxFFTIdx] = size(W);
fftLength = maxFFTIdx * 2;
% compute matrix X(fftIndex,timeFrameIndex) short term spectra
SPEC = computeSpectrum(fftLength,winShift,s);
% apply mel filter to spectra
MEL.M = W * SPEC.X;
%copy energy vector
MEL.e = SPEC.e ;
```

### B.5 Sub-function computeSpectrum.m

```matlab
function SPEC = computeSpectrum (fftLength,winShift,s)
% compute local variables
nofSamples = size(s);
maxFFTIdx = fftLength/2;
% compute time window
win = hamming(fftLength);
% compute matrix X(fftIndex,timeFrameIndex) short term spectra
k = 1;
for m = 1:winShift:nofSamples-fftLength
    spec = fft( (win.*s(m:m+fftLength-1)) ,fftLength);
    %use only lower half of fft coefficients
    SPEC.X(:,k) = ( abs( spec(1:maxFFTIdx) ) ).^2;
```

```matlab
    %compute energy
    SPEC.e(k) = sum(SPEC.X(:,k));
    k = k+1;
end
```

## B.6 Sub-function dp_asym.m

```matlab
function DP = dp_asym(X,Y)
% get number of frames for X and Y
xFrames = size(X,2);
yFrames = size(Y,2);
% create dp arrays
DP.D = zeros(yFrames,xFrames);
%create backtracking matrix
DP.B = zeros(yFrames,xFrames);
%init distance matrix for j = 1
DP.D(:,1) = Inf;
%init distance for point (1,1)
DP.D(1,1) = norm( X(:,1) - Y(:,1) );
%init backtracking matrix
DP.B(:,1) = -1;
% for all j with i = 1 only  the horizontal path
% from predecessor (i,j-1) to point (i,j) is allowed
i = 1;
for j = 2:xFrames
    DP.D(i,j) = DP.D(i,j-i) + norm(X(:,j)-Y(:,i));
    DP.B(i,j) = 0;
end
% for all j with i = 2 only  horizontal and diagonal paths
% from predecessors (i,j-1) and (i-1,j-i) to point (i,j) are allowed
i = 2;
for j = 2:xFrames
    %find best predecessor
    %horizontal path
    t = DP.D(i,j-1);
    min = t ;
    DP.B(i,j) = 0;
    %test diagonal path
    t = DP.D(i-1,j-1);
    if (t < min)
        min = t;
        DP.B(i,j) = 1 ;
    end
    DP.D(i,j) = min + norm(X(:,j)-Y(:,i));
end
%for all i > 2 all predecessor paths are allowed
for j = 2:xFrames
    for i = 3:yFrames
        %find best predecessor
        %horizontal path
        t = DP.D(i,j-1);
        min = t ;
        DP.B(i,j) = 0;
        %all others
        for p = 1:2
            t = DP.D(i-p,j-1);
            if (t < min)
                min = t;
                DP.B(i,j) = p;
            end
        end
    end
```

```matlab
            DP.D(i,j) = min + norm(X(:,j)-Y(:,i));
        end
end
%optimum path is restricted to upper-right corner of the matrix
DP.dist = DP.D(yFrames,xFrames);
%generate matching function matrix
DP.M = zeros(yFrames,xFrames);
% backtracking procedure
%init
i = yFrames;
j = xFrames;
%enter endpoint
DP.M(i,j) = 1;
back = DP.B(i,j);

%backtracking
while (back >= 0)
    %get predecessor index i
    i = i - back;
    %predecessor index j
    j = j - 1;

    %enter predecessor point into matching matrix
    DP.M(i,j) = 1;

    %get new predecessor
    back = DP.B(i,j);
end
```

## B.7 Sub-function melFilterMatrix.m

```matlab
function W = melFilterMatrix(fs, N, nofChannels)
%fs = 8000;
%N = 256;
%nofChannels = 22;
df = fs/N; %frequency resolution
Nmax = N/2; %Nyquist frequency index
fmax = fs/2; %Nyquist frequency
melmax = freq2mel(fmax); %maximum mel frequency
%mel frequency increment generating 'nofChannels' filters
melinc = melmax / (nofChannels + 1);
%vector of center frequencies on mel scale
melcenters = (1:nofChannels) .* melinc;
%vector of center frequencies [Hz]
fcenters = mel2freq(melcenters);
indexcenter = round(fcenters ./df);
%compute startfrequency, stopfrequency and bandwidth in indices
indexstart = [1 , indexcenter(1:nofChannels-1)];
indexstop = [indexcenter(2:nofChannels),Nmax];
%idxbw = (indexstop - indexstart)+1;
%compute matrix of triangle-shaped filter coefficients
W = zeros(nofChannels,Nmax);
for c = 1:nofChannels
    %left ramp
    increment = 1.0/(indexcenter(c) - indexstart(c));
    for i = indexstart(c):indexcenter(c)
        W(c,i) = (i - indexstart(c))*increment;
    end %i
    %right ramp
    decrement = 1.0/(indexstop(c) - indexcenter(c));
```

```matlab
        for i = indexcenter(c):indexstop(c)
            W(c,i) = 1.0 - ((i - indexcenter(c))*decrement);
        end %i
end %c
%normalize melfilter matrix
for j = 1:nofChannels
    W(j,:) = W(j,:)/ sum(W(j,:)) ;
end
```

## B.8 Sub-function mel2freq.m

```matlab
function b = mel2freq (m)
% compute frequency from mel value
b = 700*((10.^(m ./2595)) -1);
```

## B.9 Sub-function freq2mel.m

```matlab
function m = freq2mel (f)
% compute mel value from frequency f
m = 2595 * log10(1 + f./700);
```

## B.10 Sub-function loglimit.m

```matlab
function y = loglimit(x,limit)
% y = loglimit(x,limit)
% return log(x) or log(limit) if x < limit
if (x < limit)
    y = log(limit);
else
    y = log(x);
end;
```